\documentclass{article}
\PassOptionsToPackage{sort,numbers,compress}{natbib}
\usepackage[main,final]{neurips_2026}
\usepackage[utf8]{inputenc}
\usepackage[T1]{fontenc}
\usepackage{times,latexsym,microtype}
\usepackage{graphicx} \graphicspath{{figures/}}
\usepackage{amsmath,amssymb,mathabx,mathtools,amsthm,nicefrac}
\usepackage[colorlinks=true,citecolor=blue,bookmarks=true]{hyperref}
\usepackage[skip=3pt,font=small]{caption}
\usepackage[skip=3pt,font=small]{subcaption}
\usepackage{acronym}
\usepackage[linesnumbered,ruled,vlined]{algorithm2e}
\usepackage{wrapfig}
\usepackage[dvipsnames,svgnames,x11names,table]{xcolor}

\usepackage{enumitem}
\usepackage{xspace}
\usepackage[misc]{ifsym}
\usepackage{booktabs,tabularx,colortbl,multirow,multicol,array,makecell,tabularray}
\usepackage{siunitx}
\usepackage{pifont}
\usepackage{placeins}
\usepackage[capitalise,noabbrev,nameinlink]{cleveref} 
\crefname{section}{Sec.}{Secs.}
\Crefname{section}{Section}{Sections}
\crefname{figure}{Fig.}{Figs.}
\Crefname{figure}{Figure}{Figures}
\crefname{table}{Tab.}{Tabs.}
\Crefname{table}{Table}{Tables}
\crefname{equation}{Eq.}{Eqs.}
\Crefname{equation}{Equation}{Equations}
\crefname{algocf}{Alg.}{Algs.}
\Crefname{algocf}{Algorithm}{Algorithms}

\makeatletter
\DeclareRobustCommand\onedot{\futurelet\@let@token\@onedot}
\def\@onedot{\ifx\@let@token.\else.\null\fi\xspace}

\makeatother

\newcommand{\cmark}{\textcolor{green!70!black}{\ding{51}}}
\newcommand{\xmark}{\textcolor{red!70!black}{\ding{55}}}
\newcommand{\std}[2]{\ensuremath{#1\,{\scriptstyle\textcolor{gray}{(\pm #2)}}}}
\newcommand{\stds}[2]{\ensuremath{#1\,{\scriptscriptstyle\textcolor{gray}{(\pm #2)}}}}

\newcommand{\model}{\textbf{\textsc{LessMimic}}\xspace}

\acrodef{dof}[DoF]{Degree-of-Freedom}
\acrodef{llm}[LLM]{Large Language Model}
\acrodef{rl}[RL]{Reinforcement Learning}
\acrodef{mpc}[MPC]{Model Predictive Control}
\acrodef{mlp}[MLP]{Multi-Layer Perceptron}
\acrodef{slam}[SLAM]{Simultaneous Localization And Mapping}
\acrodef{mdp}[MDP]{Markov Decision Process}
\acrodef{ppo}[PPO]{Proximal Policy Optimization}
\acrodef{imu}[IMUs]{Inertial Measurement Units}
\acrodef{ood}[OOD]{Out-of-Distribution}
\acrodef{mocap}[MoCap]{Motion Capture}
\acrodef{sr}[SR]{Success Rate}
\acrodef{vla}[VLA]{Vision-Language-Action}
\acrodef{mpjpe}[MPJPE]{Mean Per-Joint Position Error}
\acrodef{sota}[SOTA]{State-of-the-Art}
\acrodef{fk}[FK]{Forward Kinematics}
\acrodef{ffn}[FFN]{Feed-Forward Network}
\acrodef{udp}[UDP]{User Datagram Protocol}
\acrodef{fifo}[FIFO]{First-In-First-Out}
\acrodef{gmr}[GMR]{General Motion Retargeting}
\acrodef{rfi}[RFI]{Random Force Injection}
\acrodef{com}[CoM]{Center of Mass}
\acrodef{amp}[AMP]{Adversarial Motion Priors}
\acrodef{aip}[AIP]{Adversarial Interaction Priors}
\acrodef{df}[DF]{Distance Field}
\acrodef{vae}[VAE]{Variational Auto-Encoder}

\title{\model: Versatile Humanoid--Object Interaction with Unified Distance Field Representations\vspace{-0.5em}}

\author{%
    \textbf{Yutang Lin$^{*,1,2,3,5,7,8}$ \quad
    Jieming Cui$^{*,1,2,3,5,7,8}$ \quad
    Yixuan Li$^{4,2,5}$} \\
    \textbf{Wei Liang$^{4}$ \quad
    Baoxiong Jia$^{\dagger,2,5}$ \quad
    Yixin Zhu$^{\dagger,3,1,5,6,7,8}$ \quad
    Siyuan Huang$^{\dagger,2,5}$} \\[0.5em]
    \normalfont\small$^{1}$Institute for AI, Peking University \quad $^{2}$Beijing Institute for General Artificial Intelligence (BIGAI) \\
    \normalfont\small$^{3}$School of Psychological and Cognitive Sciences, Peking University \\
    \normalfont\small$^{4}$School of Computer Science and Technology, Beijing Institute of Technology \\
    \normalfont\small$^{5}$State Key Lab of General AI \quad $^{6}$XS Vision \\
    \normalfont\small$^{7}$Beijing Key Laboratory of Behavior and Mental Health, Peking University \\
    \normalfont\small$^{8}$Embodied Intelligence Lab, PKU-Wuhan Institute for Artificial Intelligence \\
    $^{*}$Equal contribution \qquad $^{\dagger}$Corresponding author\vspace{-0.5em}%
}

\begin{document}

\maketitle

\begin{figure}[ht!]
    \centering
    \setcounter{figure}{-1}
    \captionsetup{type=figure}
    \begin{subfigure}[t]{\linewidth}
        \includegraphics[clip,trim=0cm 0cm 0cm 1cm,width=\linewidth]{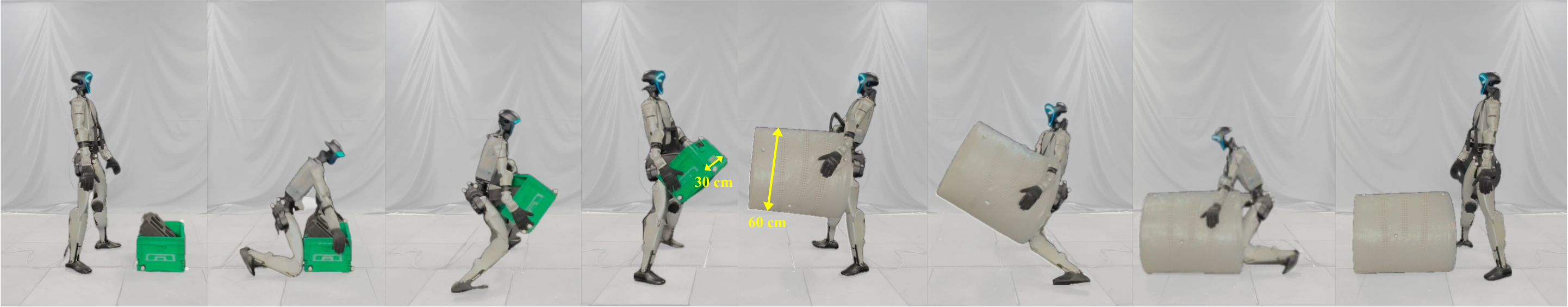}
        \caption{\textbf{Shape and size generalization.} One policy lifts objects ranging from a 30\,cm box to a 60\,cm cylinder.}
    \end{subfigure}\\
    \begin{subfigure}[t]{\linewidth}
        \includegraphics[clip,trim=0cm 0cm 0cm 0.4cm,width=\linewidth]{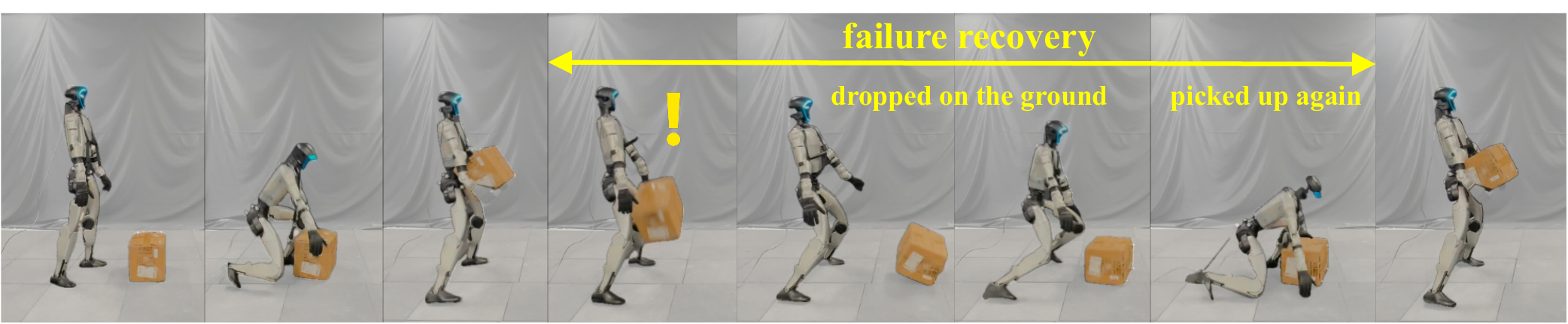}
        \caption{\textbf{Failure recovery.} After a perturbation, the humanoid restarts the pickup from the new object location.}
    \end{subfigure}\\
    \begin{subfigure}[t]{\linewidth}
        \includegraphics[clip,trim=0cm 0cm 0cm 0.5cm,width=\linewidth]{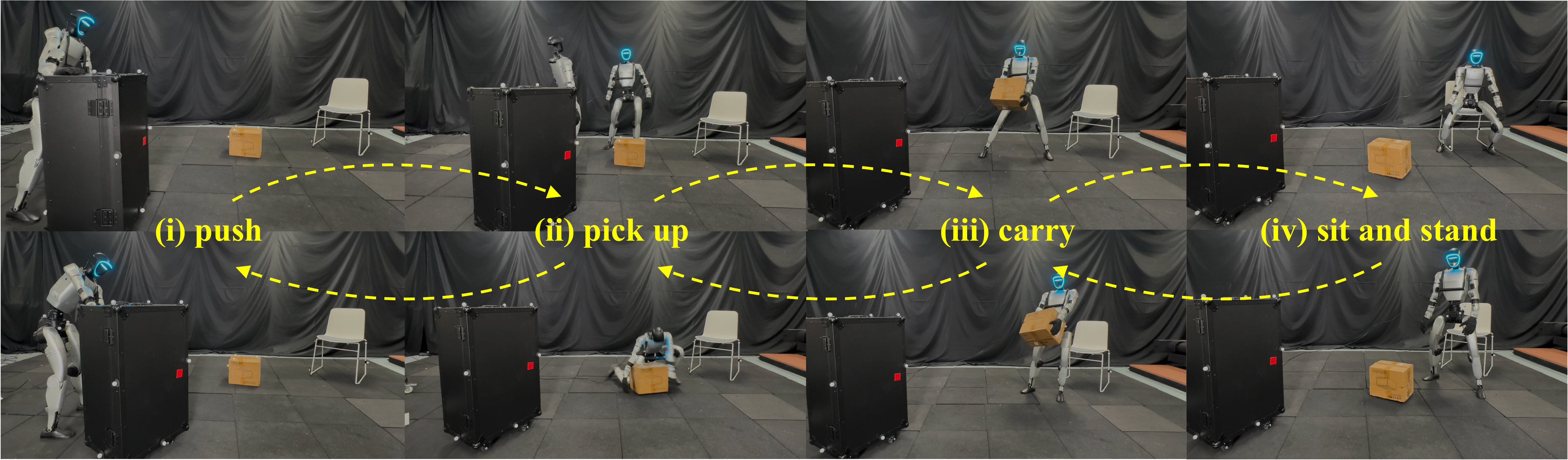}
        \caption{\textbf{Skill composition.} One policy executes Push, PickUp, Carry, and SitStand in sequence.}
    \end{subfigure}
    \captionof{figure}{\textbf{Diverse humanoid--object interaction with one \model policy.} The \acs{df}-conditioned policy (a) generalizes to unseen object shapes and scales, (b) recovers from failures through continuous geometric feedback, and (c) composes heterogeneous interaction skills.}
    \label{fig:teaser}
\end{figure}

\begin{abstract}
    Versatile whole-body interaction between humanoid robots and everyday objects is a central goal of embodied intelligence.
    Existing methods often rely on motion-reference inputs or task-specific rewards, which couple each policy to particular motion scripts, object geometries, and contact timings.
    This coupling limits both geometric generalization and flexible interaction control, in which simple commands specify the motion intent and the object geometry determines how contact is executed.
    Here we present \model, a framework for versatile humanoid--object interaction that requires no motion reference at inference. A single whole-body policy is steered by root-velocity commands and an interaction-type flag, and it is conditioned on a compact interaction representation built from short histories of surface distances and local surface directions queried from a \ac{df}. This geometry-conditioned interface lets the commands set the motion intent while contact behavior adapts to the local object geometry. Visual distillation further transfers the interface to egocentric depth, which yields a variant that deploys without \ac{mocap}.
    Across object scales from $0.4\times$ to $1.6\times$, \model performs robustly on four interaction tasks (PickUp, SitStand, Push, and Carry), whereas motion-conditioned baselines degrade sharply away from the training scale. The same policy also composes heterogeneous skills in sequence, succeeding on 62.1\% of five-task sequences and on 23.5\% of sequences with 15 task instances.
    Grounding interaction control in local geometry rather than in motion references thus offers a path toward humanoid--object control that generalizes across geometries, follows simple commands, and composes heterogeneous skills.
\end{abstract}

\acresetall
\section{Introduction}

Consider a humanoid tidying a room. It pushes a chair aside, picks up a box, carries the box across the room, and sits down to rest. Each step involves different contact patterns, object geometries, and body configurations, yet a person can perform the whole sequence without following a fixed motion script. Recent humanoid controllers achieve impressive locomotion and dynamic motion \citep{he2024omnih2o,liao2025beyond,zhang2025track,li2025bfm,li2025clone,du2025learning}, but reproducing this versatility with varied objects on hardware remains an open problem in embodied intelligence \citep{nakamura1987task,khatib2004whole,dietrich2015overview}.

The core obstacle is representational. Current approaches differ mainly in whether the deployed policy requires motion references, and both choices impose limitations. Motion-conditioned methods \citep{yang2025omniretarget,weng2025hdmi,zhao2025resmimic,liao2025beyond,zhang2025track} condition the policy on recorded demonstrations and achieve high-fidelity motion. This formulation, however, entangles object geometry, contact timing, and body motion with specific reference trajectories. The policy can then become a geometric specialist that loses flexible interaction control, because every deviation from the reference is penalized as a tracking error, even when the deviation is needed to adapt to a new object, a changed root command, or accumulated state drift. Recent variants add higher-level planners or human-in-the-loop teleoperation \citep{xue2025leverb,luo2025sonic,fu2025demohlm,yin2025visual,li2025clone,ze2025twist2}, but these mainly compensate for the same representational bottleneck, at the cost of additional perceptual complexity or human involvement. Motion-free-at-inference methods remove motion references from the deployed policy and can be steered more flexibly, but without a principled interaction signal they often rely on task-specific rewards or handcrafted observations \citep{wang2025physhsi,haarnoja2024learning}, which yields isolated policies that do not naturally compose across heterogeneous skills. What is missing is a geometry-aware representation, shared across tasks, that decouples interaction logic from specific motion patterns, so that simple commands specify the motion intent and the local geometry determines how contact is executed.

We argue that the \ac{df} \citep{osher1988fronts} provides a natural basis for this representation. A \ac{df} maps each query point to its distance from the nearest object surface and thereby gives compact proximity information and, where the field is differentiable, local surface-direction cues. Near non-smooth regions, numerical direction estimates still provide practical contact-relevant cues, so the representation does not require global differentiability. Compared with point clouds, voxels, or implicit neural geometry, \acp{df} offer efficient, geometry-aligned queries suited to high-frequency whole-body control. Expressed as local distances and surface directions, and combined with decomposed link velocities during post-training, \acp{df} supply the interaction signal for geometry-aware contact adaptation while reducing the dependence on absolute object coordinates and demonstration-specific trajectories.

We introduce \model, a framework for versatile humanoid--object interaction that is motion-free at inference, meaning that the deployed policy consumes no reference joint poses, future body motion, or contact trajectories. A root-velocity command and an interaction-type flag steer the policy, while local \ac{df} observations ground its contact behavior in the current object geometry. \model encodes short histories of \ac{df}-derived surface distances and directions into a compact interaction representation and trains a single whole-body policy in three stages. Teacher-guided behavior cloning gives a stable, contact-rich initialization, \ac{rl} under randomized object geometries uses \acf{aip} as a cross-task interaction reward, and visual distillation approximates the geometric interface from egocentric depth.

The result is one low-level policy interface for command-steered humanoid--object control (\cref{fig:teaser}). \model uses this interface unchanged across four tasks (PickUp, SitStand, Push, and Carry) and object scales from $0.4\times$ to $1.6\times$; on PickUp at $0.4\times$, it succeeds in 63.0\% of episodes, against 0.0\% for our reproductions of the motion-conditioned HDMI \citep{weng2025hdmi} and ResMimic \citep{zhao2025resmimic}. In sequential multi-skill evaluation, the same policy executes externally commanded interaction modes without skill-specific policies and succeeds on 62.1\% of five-task sequences and on 23.5\% of sequences with 15 task instances, whereas most ablated variants collapse.

Our contributions are threefold.
\begin{itemize}[nosep,leftmargin=*]
    \item \textbf{A unified \acs{df}-based interaction representation.} Its geometry-aware robot--object cues let one policy adapt its behavior to object scale, shape, and surface geometry.
    \item \textbf{A learning pipeline that is motion-free at inference.} It transfers physically valid contact behavior from a motion-imitation teacher into a command-steered policy and improves generalization under randomized object geometries with \ac{aip}-guided \ac{rl}.
    \item \textbf{A visual distillation scheme.} It transfers the \ac{df}-based interface from privileged object geometry to egocentric depth, which yields a variant that deploys without \ac{mocap}.
\end{itemize}

\section{Related Work}

\subsection{Motion-Conditioned Humanoid--Object Interaction}

\begin{wraptable}{r}{0.48\linewidth}
    \vspace{-8pt}
    \centering
    \small
    \setlength{\tabcolsep}{3pt}
    \caption{\textbf{Comparison of humanoid--object interaction methods.} Check marks indicate inference without motion or pose references, a unified geometric observation, and multi-skill learning in one policy.}
    \label{tab:interaction_comparison}
    \resizebox{\linewidth}{!}{%
        \begin{tabular}{lccc}
            \toprule
            \textbf{Method} & \textbf{\makecell{No motion/pose\\ref. input}} & \textbf{\makecell{Unified geometric\\observation}} & \textbf{\makecell{One-policy\\multi-skill learning}} \\
            \midrule
            HDMI \citep{weng2025hdmi} & \xmark & \xmark & \cmark \\
            ResMimic \citep{zhao2025resmimic} & \xmark & \xmark & \cmark \\
            CLONE \citep{li2025clone} & \xmark & \xmark & \cmark \\
            HumanX \citep{wang2026humanx} & \cmark & \xmark & \xmark \\
            HAIC \citep{li2026haic} & \xmark & \cmark & \xmark \\
            PhysHSI \citep{wang2025physhsi} & \cmark & \xmark & \xmark \\
            \midrule
            \textbf{Ours} & \cmark & \cmark & \cmark \\
            \bottomrule
        \end{tabular}%
    }%
\end{wraptable}

\Ac{rl} has enabled increasingly capable whole-body humanoid control for locomotion, dynamic motion, and physical interaction \citep{moro2019whole,fu2024humanplus,he2024omnih2o,ji2024exbody2,wang2025experts,xie2025kungfubot,shi2025adversarial}. A dominant paradigm conditions humanoid--object policies on reference motions or target poses to stabilize learning and produce plausible behavior \citep{weng2025hdmi,liao2025beyond,zhang2025track,zhao2025resmimic,yin2025visual,allshire2025visual,xue2025leverb}. Such motion-centric formulations tie contact timing, object geometry, and motion topology to the reference data, which limits generalization when object shape, scale, or configuration changes.

Other systems broaden skill acquisition through scalable data generation, teleoperation, or human-in-the-loop correction \citep{jing2025humanoidgen,li2025clone,ze2025twist2,ze2025twist,luo2025sonic,li2025amo,he2024omnih2o,fu2024humanplus,he2024learning,wang2026humanx}. They increase the availability and diversity of motion supervision but provide no unified geometric policy interface that serves heterogeneous interaction skills without motion or pose references.

\Cref{tab:interaction_comparison} summarizes these distinctions. HDMI and ResMimic learn multiple skills within one policy, but their deployed interface remains reference-conditioned, so composing a new heterogeneous sequence requires corresponding motion and object references, including the transitions between skills. CLONE supports sequential interaction through teleoperation. HumanX learns a broad range of humanoid skills from human videos, but it targets scalable skill acquisition rather than a steerable low-level interaction interface. HAIC \citep{li2026haic} uses geometric observations yet remains motion-conditioned at inference, and PhysHSI is motion-free at inference but uses no unified geometric observation. \model is motion-free at inference, uses a unified \ac{df}-based geometric observation, and switches between heterogeneous skills online from the interaction-type flag and root commands, so no full-body reference has to be collected for each composed sequence.

\subsection{Motion-Free-at-Inference Humanoid--Object Interaction}

Removing motion or pose references at inference is important for flexible humanoid interaction control \citep{lunardi2023reference}. Classical model-based control and online optimization generate behavior from objectives without prerecorded motion references \citep{posa2013direct,schulman2014motion,herzog2016momentum,sentis2005synthesis,hopkins2016optimization}, but they often require accurate models and carefully designed objectives. Learning-based methods instead condition policies on task-relevant observations \citep{li2025bfm,li2025hold,wang2025physhsi,peng2021amp,tessler2023calm,zhang2023efficient,kumar2021rma,haarnoja2024learning}; PhysHSI \citep{wang2025physhsi}, for example, studies motion-free humanoid--scene interaction. These methods reduce the reliance on deployed motion references but often remain tied to task-specific observations, rewards, dynamics assumptions, or specialized domains. \model uses one unified geometric observation across heterogeneous humanoid--object interaction skills.

\subsection{Geometric Representations for Interaction}

Interaction methods encode geometry as voxels and occupancy maps \citep{liu2025fetchbot,huang2023voxposer,varley2017shape,xu2020learning,jiang2021synergies}, points or meshes \citep{zhu2024point,jia2025pointmappolicy,chisari2024learning,liu2022frame,zesch2023neural,shen2021igibson,pan2012fcl,hauser2016robust}, or implicit neural representations \citep{song2025implicit,yang2025contactsdf,quintero2024stochastic,dai2022graspnerf}. These representations trade off geometric detail, query efficiency, and suitability for high-frequency contact-aware control. \Acp{df} have appeared in robotics, for example to represent articulated robot geometry \citep{li2024rdf}, and a humanoid whole-body policy for embracing bulky objects \citep{zheng2025bulky} learns from object-level cues such as the distance to the object center. Prior work, however, has not used object-surface distance and direction queries at humanoid body links as a deployed policy observation for versatile humanoid--object interaction. \model queries the \ac{df} at a fixed set of humanoid links, so each query relates one link to the object surface rather than to the object as a whole. These queries form one unified interaction representation across heterogeneous skills, and \ac{aip} further regularizes the resulting interaction statistics.

\begin{figure}[t!]
    \centering
    \includegraphics[width=\linewidth]{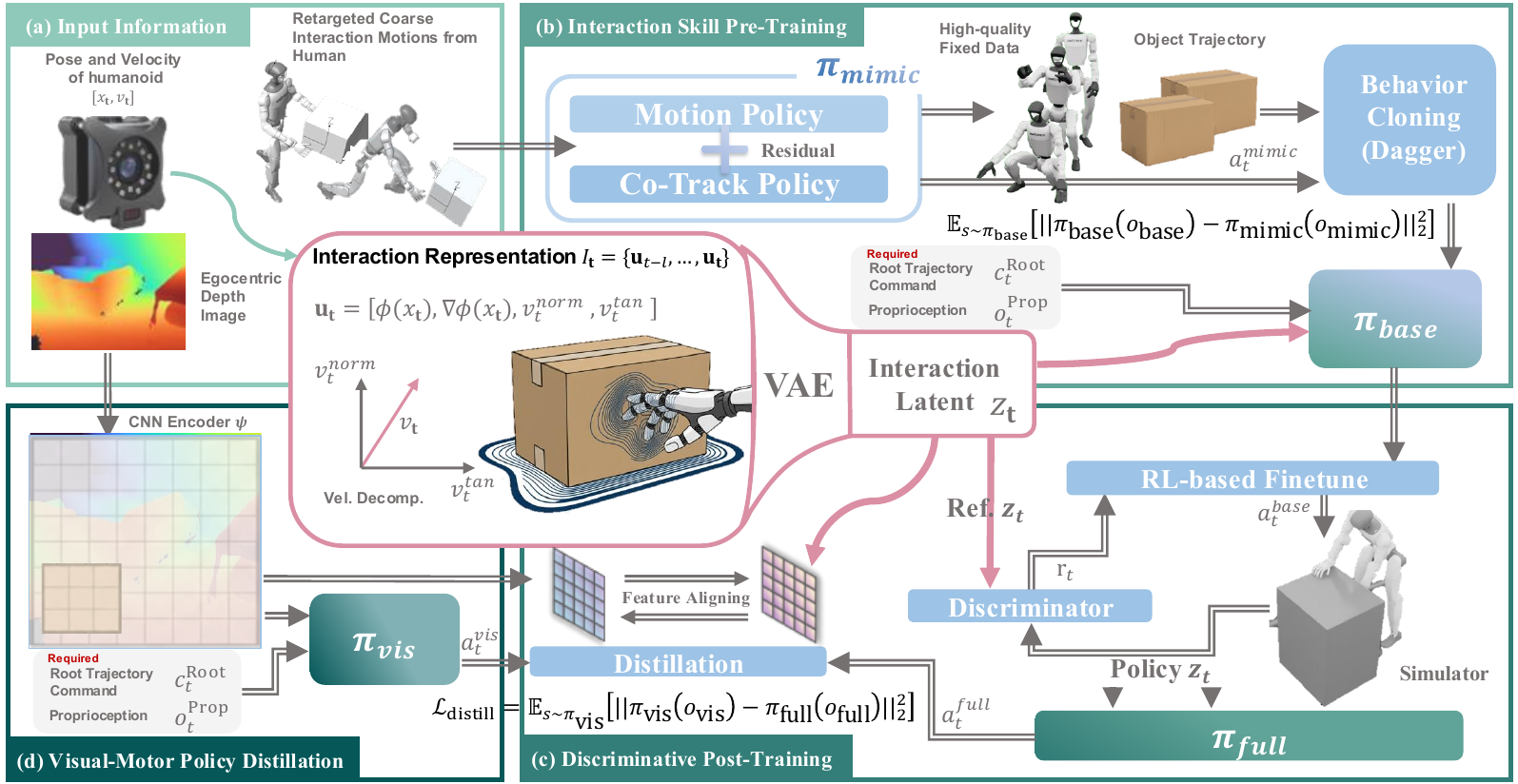}
    \caption{\textbf{Training pipeline of \model.} (a) \acs{df}-based robot--object cues from \acs{mocap} or egocentric depth are encoded into an interaction latent $z_t$. (b) A motion-tracking teacher $\pi_{\mathrm{mimic}}$ provides physically feasible demonstrations for behavior cloning of $\pi_{\mathrm{base}}$, which uses the inference-time observations. (c) $\pi_{\mathrm{base}}$ is fine-tuned with \acs{aip}-guided \acs{rl} under randomized geometries into $\pi_{\mathrm{full}}$. (d) $\pi_{\mathrm{full}}$ is distilled into a vision policy $\pi_{\mathrm{vis}}$ for deployment without \acs{mocap}.}
    \label{fig:pipeline}
\end{figure}

\section{The \model Framework}\label{sec:method}

\model couples a \ac{df}-based interaction representation with a whole-body controller that is motion-free at inference (\cref{fig:pipeline}). This coupling supports scale generalization and the sequential composition of heterogeneous interaction skills. \Cref{sec:method:interaction} defines the representation, and the three training stages follow in order, from interaction skill pre-training (\cref{sec:method:pretrain}) through discriminative post-training (\cref{sec:method:posttrain}) to visual-motor distillation (\cref{sec:method:distill}).

\subsection{\texorpdfstring{\acs{df}}{DF}-Based Interaction Representation}\label{sec:method:interaction}

\model describes interaction through local robot--object geometry rather than absolute motion trajectories. Let $\Phi:\mathbb{R}^3\rightarrow\mathbb{R}_{\ge0}$ be the unsigned Euclidean distance to the object surface. The \ac{df} is queried at a fixed set $\mathcal{K}$ of humanoid links consisting of both wrists, both ankles, and the pelvis. At time $t$, the query at the position $\mathbf{x}_{t,k}$ of each link $k\in\mathcal{K}$ returns the distance $d_{t,k}=\Phi(\mathbf{x}_{t,k})$ and a local surface-direction estimate $\mathbf{n}_{t,k}$. Where $\Phi$ is differentiable, $\mathbf{n}_{t,k}=\nabla\Phi(\mathbf{x}_{t,k})/\|\nabla\Phi(\mathbf{x}_{t,k})\|_2$; near non-smooth regions, such as edges or points with non-unique projections, a normalized numerical estimate replaces the gradient. The method therefore relies on stable local direction cues rather than on global differentiability of the \ac{df}. A batched pass over these queries takes less than $9\,\mathrm{ms}$ on one RTX 5090, even for 1024 parallel environments with three objects (\cref{tab:supp_df_latency}).

Each link contributes the feature
\begin{equation}
    \mathbf{u}_{t,k}=[d_{t,k},\mathbf{n}_{t,k}],
\end{equation}
and the interaction history over a window of $l=5$ steps is
\begin{equation}
    I_t=\{\mathbf{u}_{\tau,k}\}_{\tau=t-l+1:t,\;k\in\mathcal{K}}.
\end{equation}
This history is not a complete reconstruction of the object geometry but a summary of the contact-relevant local proximity and surface-direction cues for whole-body control. Because it is defined by local robot--object relations rather than by absolute object coordinates or reference trajectories, it reduces the dependence on object-specific demonstrations.

A \ac{vae} encodes $I_t$ into a compact 64-dimensional latent $z_t$, which gives the policy a fixed-size interaction representation and smooths noisy or discontinuous local \ac{df} queries (\cref{sec:supp:method:interaction}). During discriminative post-training, \ac{aip} augments this latent with link velocities expressed in the local surface frame (\cref{sec:method:posttrain}).

\subsection{Interaction Skill Pre-Training}\label{sec:method:pretrain}

Learning contact-rich whole-body interaction directly from sparse objectives is unstable, because meaningful rewards often appear only after balance, locomotion, and object contact are already coordinated. We therefore use a physics-based mimic teacher, but only as training supervision. Following ResMimic \citep{zhao2025resmimic}, the teacher $\pi_{\mathrm{mimic}}$ is a single multi-task policy that tracks multiple retargeted human--object interaction motions associated with three training object instances and the four task categories PickUp, SitStand, Push, and Carry. It receives the current reference-motion state together with privileged interaction observations such as body velocity, object pose, and contact, and produces feasible contact behavior under full simulation physics.

The student policy $\pi_{\mathrm{base}}$ is trained on the observation used at inference,
\begin{equation}
    o_{\mathrm{base}}=[o_{\mathrm{prop}},c^{\mathrm{root}}_t,g_t,z_t],
    \label{eq:obs_base}
\end{equation}
where $o_{\mathrm{prop}}$ is proprioception, $c^{\mathrm{root}}_t\in\mathbb{R}^2$ is the commanded planar root velocity, $g_t$ is a one-hot interaction-type flag, and $z_t$ is the \ac{df}-based interaction latent. The flag specifies the intended skill mode but carries no reference joint poses, future body motion, or contact trajectories. All three stages use the same Transformer-based whole-body controller architecture, whose input tokens encode these four components and whose output is the whole-body action (\cref{sec:supp:method:training}).

We train $\pi_{\mathrm{base}}$ with DAgger-style behavior cloning \citep{ross2011reduction}, in which the student acts in simulation and the teacher provides corrective actions at the visited states. The objective is
\begin{equation}
    \mathcal{L}_{\mathrm{BC}}=\mathbb{E}_{s\sim\pi_{\mathrm{base}}}\left[\left\|\pi_{\mathrm{base}}(o_{\mathrm{base}})-\pi_{\mathrm{mimic}}(o_{\mathrm{mimic}})\right\|_2^2\right],
    \label{eq:bc}
\end{equation}
where $o_{\mathrm{mimic}}$ contains privileged reference-motion information unavailable to the deployed policy. Teacher and student run in parallel simulation environments, each pairing a training object with its reference motion or motions, and rollouts from all tasks and objects are sampled jointly to update the single $\pi_{\mathrm{base}}$. This stage applies no object-scale randomization; geometry randomization enters only in discriminative post-training. Pre-training thus transfers contact-rich imitation supervision into a policy that is motion-free at inference.

\subsection{Discriminative Post-Training}\label{sec:method:posttrain}

Pre-training provides physically feasible contact behavior, but the policy can still inherit object-specific biases from the teacher trajectories. We therefore fine-tune $\pi_{\mathrm{base}}$ with \ac{rl} under randomized object geometries, varying scale, shape, and surface properties. The reward combines command tracking, whole-body control regularization, and the proposed interaction prior. Motion-reference tracking is removed deliberately \emph{before} geometry-randomized \ac{rl}, so that the policy learns to adapt to changed geometry through the same \ac{df}-conditioned interface it uses at deployment.

We introduce \acf{aip} as a cross-task interaction reward computed on velocity-augmented \ac{df} interaction statistics. It compares local robot--object interaction patterns instead of regularizing full-body kinematic trajectories, because valid interactions with different geometries may require different joint configurations yet should preserve similar proximity, approach, and surface-traversal behavior. For each queried link $k$, the link velocity $\mathbf{v}_{t,k}$ is decomposed in the local surface frame as
\begin{equation}
    \mathbf{v}^{\mathrm{norm}}_{t,k}=(\mathbf{v}_{t,k}^{\top}\mathbf{n}_{t,k})\mathbf{n}_{t,k},\qquad\mathbf{v}^{\mathrm{tan}}_{t,k}=\mathbf{v}_{t,k}-\mathbf{v}^{\mathrm{norm}}_{t,k},
\end{equation}
and the \ac{aip} discriminator receives
\begin{equation}
    h_t=[z_t,\{\mathbf{v}^{\mathrm{norm}}_{t,k},\mathbf{v}^{\mathrm{tan}}_{t,k}\}_{k\in\mathcal{K}}],
\end{equation}
where $z_t$ is the \ac{df}-based latent used by the policy. The explicit velocities are needed because the temporal \ac{df} history captures changes in distance and surface direction but does not uniquely identify tangential motion; on a locally planar surface, static contact and sliding at different tangential velocities can produce nearly identical distance and normal histories. These velocities enter only the \ac{aip} discriminator during post-training; the deployed policy remains conditioned on $z_t$.

We build a reference buffer $\mathcal{B}_{\mathrm{ref}}$ from rollouts of the mimic teacher $\pi_{\mathrm{mimic}}$ and train a discriminator $D$ to separate its samples from policy samples with
\begin{equation}
    \mathcal{L}_D=\mathbb{E}_{h\sim\mathcal{B}_{\mathrm{ref}}}\left[(D(h)-1)^2\right]+\mathbb{E}_{h\sim\pi}\left[(D(h)+1)^2\right].
\end{equation}
The discriminator then defines the interaction reward
\begin{equation}
    r_{\mathrm{interact}}(h_t)=\max\!\left(0,\,1-0.25(D(h_t)-1)^2\right).
    \label{eq:interaction_reward}
\end{equation}
The complete post-training reward is
\begin{equation}
    r_t=r_{\mathrm{cmd}}+\lambda_i r_{\mathrm{interact}}(h_t)+\lambda_s r_{\mathrm{style}}+r_{\mathrm{reg}},
    \label{eq:posttrain_reward}
\end{equation}
where $r_{\mathrm{cmd}}$ tracks the root command, $r_{\mathrm{style}}$ regularizes motion naturalness, $r_{\mathrm{reg}}$ collects standard control and safety regularizers, and $\lambda_i$ and $\lambda_s$ weight the interaction and style rewards (\cref{sec:supp:method:posttrain}). Because the discriminator sees compact interaction statistics rather than the full robot state, \ac{aip} encourages geometrically valid interaction without enforcing a specific motion trajectory. Post-training yields the policy $\pi_{\mathrm{full}}$.

\begin{table}[t!]
    \centering
    \small
    \setlength{\tabcolsep}{3pt}
    \caption{\textbf{Object-scale generalization in simulation.} Column headers give object scale factors; $0.4\times$ and $1.6\times$ lie outside our per-axis post-training range $[0.6,1.4]$. Ablations remove \acs{aip}, synthetic interaction data, geometry randomization, \acs{rl}, or the Transformer. $R_{\mathrm{succ}}$ is the task success rate and $R_{\mathrm{cont}}$ the hand--object contact rate; entries are mean $\pm$ std over three seeds. Bold and underline mark the best and second-best results, $^*$ our implementation, $^\dagger$ methods with one policy per task, and dashes settings that are not reported.}
    \label{tab:unified_sim}
    \resizebox{\linewidth}{!}{%
        \begin{tabular}{lccccc ccc ccc c}
            \toprule
            \textbf{Method} & \multicolumn{5}{c}{\textbf{PickUp $R_{\mathrm{succ}}$ ($\uparrow$, \%)}} & \multicolumn{3}{c}{\textbf{SitStand $R_{\mathrm{succ}}$ ($\uparrow$, \%)}} & \multicolumn{3}{c}{\textbf{Push $R_{\mathrm{cont}}$ ($\uparrow$, \%)}} & \multicolumn{1}{c}{\textbf{Carry $R_{\mathrm{succ}}$ ($\uparrow$, \%)}} \\
            \cmidrule(lr){2-6}\cmidrule(lr){7-9}\cmidrule(lr){10-12}\cmidrule(lr){13-13}
            & \textbf{0.4} & \textbf{0.6} & \textbf{1.0} & \textbf{1.4} & \textbf{1.6} & \textbf{0.4} & \textbf{1.0} & \textbf{1.6} & \textbf{0.6} & \textbf{1.0} & \textbf{1.4} & \textbf{0.4--1.6 Avg.} \\
            \midrule
            \rowcolor{gray!30}
            \multicolumn{13}{l}{\textbf{Motion-based baselines}} \\
            HDMI$^*$ & \std{0.0}{0.0} & \std{\underline{99.7}}{0.5} & \std{\textbf{100.0}}{0.0} & \std{40.7}{3.3} & \std{1.7}{1.2} & \std{0.0}{0.0} & \std{\underline{99.0}}{0.0} & \std{1.7}{0.5} & \std{0.3}{0.0} & \std{\textbf{97.3}}{0.5} & \std{\textbf{10.6}}{2.9} & \std{27.4}{0.3} \\
            ResMimic$^*$ & \std{0.0}{0.0} & \std{18.3}{1.2} & \std{\textbf{100.0}}{0.0} & \std{\textbf{99.7}}{0.5} & \std{63.0}{10.2} & \std{0.0}{0.0} & \std{\textbf{100.0}}{0.0} & \std{93.7}{0.5} & \std{0.4}{0.2} & \std{\underline{93.1}}{1.2} & \std{2.2}{0.7} & \std{32.9}{1.5} \\
            VisualMimic$^\dagger$ & \std{0.0}{0.0} & \std{0.0}{0.0} & \std{\textbf{100.0}}{0.0} & \std{0.0}{0.0} & \std{0.0}{0.0} & -- & -- & -- & -- & \std{20.6}{9.0} & -- & -- \\
            \midrule
            \rowcolor{gray!30}
            \multicolumn{13}{l}{\textbf{Motion-free-at-inference baselines}} \\
            PhysHSI$^\dagger$ & \std{23.1}{2.4} & \std{70.5}{2.6} & \std{\textbf{100.0}}{0.0} & \std{47.9}{2.8} & \std{39.9}{2.8} & \std{61.3}{2.7} & \std{76.2}{2.3} & \std{71.8}{2.4} & -- & -- & -- & \std{\underline{81.6}}{1.9} \\
            \midrule
            \rowcolor{gray!30}
            \multicolumn{13}{l}{\textbf{Ablation study}} \\
            Ours - AIP & \std{0.0}{0.0} & \std{0.7}{0.5} & \std{23.3}{5.0} & \std{57.3}{1.2} & \std{64.0}{2.9} & \std{\textbf{98.7}}{0.5} & \std{98.3}{0.5} & \std{89.0}{0.8} & \std{\textbf{82.9}}{1.1} & \std{72.5}{1.9} & \std{\underline{6.5}}{1.9} & \std{0.0}{0.0} \\
            Ours - Syn. & \std{34.0}{6.7} & \std{94.3}{2.5} & \std{\underline{99.3}}{0.9} & \std{\textbf{99.7}}{0.5} & \std{\textbf{99.7}}{0.5} & \std{95.3}{1.7} & \std{30.0}{2.2} & \std{86.3}{3.9} & \std{0.0}{0.0} & \std{41.9}{2.1} & \std{3.5}{1.6} & \std{66.5}{2.0} \\
            Ours - Rand. & \std{0.0}{0.0} & \std{0.0}{0.0} & \std{0.0}{0.0} & \std{0.0}{0.0} & \std{0.0}{0.0} & \std{0.0}{0.0} & \std{81.3}{3.4} & \std{\textbf{97.7}}{1.2} & \std{0.0}{0.0} & \std{67.0}{1.6} & \std{2.1}{1.0} & \std{0.0}{0.0} \\
            Ours - RL & \std{0.0}{0.0} & \std{2.0}{0.8} & \std{31.7}{2.6} & \std{37.0}{2.4} & \std{9.7}{1.7} & \std{1.7}{1.2} & \std{64.7}{3.1} & \std{15.0}{2.8} & \std{0.0}{0.0} & \std{67.3}{0.7} & \std{0.0}{0.0} & \std{5.3}{1.5} \\
            Ours - Trans. & \std{0.0}{0.0} & \std{0.0}{0.0} & \std{0.0}{0.0} & \std{0.0}{0.0} & \std{0.0}{0.0} & \std{\underline{98.0}}{0.8} & \std{77.7}{0.5} & \std{\underline{95.7}}{2.1} & \std{\underline{18.3}}{1.4} & \std{85.7}{2.4} & \std{3.0}{0.6} & \std{0.0}{0.0} \\
            \midrule
            Ours (MoCap) & \std{\underline{63.0}}{5.0} & \std{\textbf{100.0}}{0.0} & \std{\textbf{100.0}}{0.0} & \std{\underline{88.0}}{1.6} & \std{\underline{94.0}}{1.6} & \std{79.0}{0.0} & \std{80.3}{2.1} & \std{61.7}{1.2} & \std{13.9}{0.2} & \std{51.3}{1.6} & \std{2.0}{0.3} & \std{\textbf{82.9}}{1.4} \\
            Ours (Vision) & \std{\textbf{63.7}}{3.1} & \std{94.7}{1.9} & \std{91.0}{3.3} & \std{\textbf{99.7}}{0.5} & \std{93.0}{1.6} & -- & -- & -- & \std{1.2}{0.1} & \std{35.2}{1.3} & \std{1.9}{0.8} & \std{35.8}{2.7} \\
            \bottomrule
        \end{tabular}%
    }%
\end{table}

\paragraph{Root commands and synthetic command augmentation.}
Post-training draws root-velocity commands from recorded and synthetic sources. For a recorded trajectory, the training command is the finite difference of the root trajectory projected onto the $x$--$y$ plane. The 44 collected interaction clips contain limited locomotion diversity, so we augment them with synthetic samples that compose coarse root and object command primitives through predefined task templates. The primitives come from the same family used by the evaluation protocol, but the samples reuse none of the held-out composed evaluation sequences. Because inference is reference-free, the augmentation operates on commands and does not require synthesizing a new full-body motion trajectory for each command composition.

\subsection{Visual-Motor Policy Distillation}\label{sec:method:distill}

The post-trained policy $\pi_{\mathrm{full}}$ computes \ac{df} queries from \ac{mocap}-derived object geometry, which is convenient for training and controlled evaluation but limits deployment in uninstrumented environments. To relax this sensing requirement, we distill $\pi_{\mathrm{full}}$ into a vision-based policy $\pi_{\mathrm{vis}}$ that observes egocentric depth.

The vision policy keeps the control interface of $\pi_{\mathrm{full}}$ but replaces the \ac{df} latent with an interaction latent predicted from depth,
\begin{equation}
    o_{\mathrm{vis}}=[o_{\mathrm{prop}},c^{\mathrm{root}}_t,g_t,\hat z_t],\qquad\hat z_t=E_\phi(S_t),
    \label{eq:obs_vis}
\end{equation}
where $S_t$ is a history of egocentric depth frames and $E_\phi$ is a visual encoder trained to produce a compact interaction latent sufficient for the control policy. Visual distillation thus transfers the \ac{df}-based interface from privileged geometry to onboard perception.

Distillation is again DAgger-style, with the frozen $\pi_{\mathrm{full}}$ as the teacher. At each iteration, $\pi_{\mathrm{vis}}$ acts in simulation, $\pi_{\mathrm{full}}$ is queried at the visited states with its \ac{df}-based observation, and the student regresses the teacher's actions,
\begin{equation}
    \mathcal{L}_{\mathrm{distill}}=\mathbb{E}_{s\sim\pi_{\mathrm{vis}}}\left[\left\|\pi_{\mathrm{vis}}(o_{\mathrm{vis}})-\pi_{\mathrm{full}}(o_{\mathrm{base}})\right\|_2^2\right].
    \label{eq:distill}
\end{equation}
We randomize camera extrinsics, depth noise, and physical parameters to improve robustness (\cref{sec:supp:method:distill}). \Cref{sec:exp_vis} examines the performance of $\pi_{\mathrm{vis}}$ and the observability of contacts.

\section{Experiments}\label{sec:exp}

The experiments test whether \model generalizes across object scale on all four skills and across shape on PickUp (\cref{sec:exp_gen}), executes commanded multi-skill sequences without skill-specific policies (\cref{sec:exp_long}), and retains its performance when egocentric depth replaces the \ac{df} observation (\cref{sec:exp_vis}). Component ablations (\cref{sec:exp_abla}) and hardware experiments (\cref{sec:exp_real}) complete the evaluation.

\subsection{Experimental Setup}\label{sec:exp:setup}

\paragraph{Tasks and metrics.}
We evaluate four humanoid--object interaction tasks. PickUp requires lifting an object and holding it stably, SitStand requires stable pelvis--chair contact within a valid root-height range, and Carry requires transporting an object along a commanded trajectory; for these tasks we report the success rate $R_{\mathrm{succ}}$. For Push, the single-task results report the hand--object contact rate $R_{\mathrm{cont}}$, which measures contact maintenance rather than full pushing success. In sequential evaluation, every task instance, including Push, is judged by its maximum deviation from the commanded trajectory. Because the metrics differ across tasks, we report per-task results rather than one aggregated score (\cref{sec:supp:exp:metrics}).

\paragraph{Policy interface.}
At inference, the policy receives proprioception, the planar root-velocity command, the one-hot interaction-type flag, and either the \ac{df} latent or the depth-predicted latent, but no reference joint poses, future body motion, or contact trajectories. In sequential evaluation, the task order, root command, and interaction-type flag are specified externally at skill boundaries, and the policy must execute the commanded skills without environment resets or skill-specific policies.

\paragraph{Implementation and baselines.}
\Ac{df} features are queried at the fixed link set of \cref{sec:method:interaction} and encoded over a short temporal history (\cref{sec:supp:method:interaction}). We compare against the motion-conditioned baselines HDMI \citep{weng2025hdmi}, ResMimic \citep{zhao2025resmimic}, and VisualMimic \citep{yin2025visual} and against the motion-free-at-inference baseline PhysHSI \citep{wang2025physhsi}, all evaluated under the same task definitions and object-scale settings where applicable. Our post-training scales each object axis independently within $[0.6,1.4]$, the same range used by our HDMI and ResMimic reproductions, so $0.4\times$ and $1.6\times$ are strict scale extrapolations. Each policy is evaluated on 128 episodes per task and scale (\cref{sec:supp:exp:sim}).

\begin{table}[t!]
    \centering
    \small
    \setlength{\tabcolsep}{3pt}
    \caption{\textbf{Quantitative real-world deployment.} Success counts are over ten trials per setting. $R_{\mathrm{acc}}$ is the root-tracking accuracy, measured by external \acs{mocap} for evaluation only. Dashes mark SitStand, where the vision policy is not evaluated because its front camera poorly observes the back-side chair contact.}
    \label{tab:deploy}
    \resizebox{\linewidth}{!}{%
        \begin{tabular}{lcc cc cc cc}
            \toprule
            \textbf{Method} & \multicolumn{2}{c}{\textbf{PickUp $22\,\mathrm{cm}$} ($\uparrow$, \%)} & \multicolumn{2}{c}{\textbf{PickUp $60\,\mathrm{cm}$} ($\uparrow$, \%)} & \multicolumn{2}{c}{\textbf{SitStand $12\,\mathrm{cm}$} ($\uparrow$, \%)} & \multicolumn{2}{c}{\textbf{SitStand $46\,\mathrm{cm}$} ($\uparrow$, \%)} \\
            \cmidrule(lr){2-3}\cmidrule(lr){4-5}\cmidrule(lr){6-7}\cmidrule(lr){8-9}
            & \textbf{Real $R_{\mathrm{succ}}$} & \textbf{Root $R_{\mathrm{acc}}$} & \textbf{Real $R_{\mathrm{succ}}$} & \textbf{Root $R_{\mathrm{acc}}$} & \textbf{Real $R_{\mathrm{succ}}$} & \textbf{Root $R_{\mathrm{acc}}$} & \textbf{Real $R_{\mathrm{succ}}$} & \textbf{Root $R_{\mathrm{acc}}$} \\
            \midrule
            Ours (MoCap) & 10 / 10 & 94.44 & 8 / 10 & 81.39 & 8 / 10 & 84.89 & 10 / 10 & 91.88 \\
            Ours (Vision) & 8 / 10 & 89.15 & 7 / 10 & 75.24 & -- & -- & -- & -- \\
            \bottomrule
        \end{tabular}%
    }%
\end{table}

\begin{table}[t]
    \centering
    \small
    \caption{\textbf{Matched geometric-representation ablation on PickUp.} Success rates (\%) with the policy architecture, training pipeline, and scale-randomization protocol held fixed. Latent \acs{df} is the proposed temporally encoded representation; raw \acs{df} feeds the unencoded features to the policy but keeps the latent input for \acs{aip}. Bold marks the best value in each column.}
    \label{tab:repr_ablation}
    \begin{tabular*}{\linewidth}{@{\extracolsep{\fill}}lrrrrrr@{}}
        \toprule
        \textbf{Representation} & \textbf{0.4$\times$} & \textbf{0.6$\times$} & \textbf{1.0$\times$} & \textbf{1.4$\times$} & \textbf{1.6$\times$} & \textbf{Overall} \\
        \midrule
        Latent DF & \textbf{63.0} & \textbf{100.0} & \textbf{100.0} & \textbf{88.0} & \textbf{94.0} & \textbf{89.0} \\
        BBox & 14.8 & 40.6 & 48.4 & 76.6 & 82.0 & 52.5 \\
        Voxel ($16^3$) & 44.5 & 50.1 & 75.8 & 76.1 & 85.9 & 66.5 \\
        Raw DF & 56.3 & 82.8 & 94.5 & 80.1 & 92.5 & 81.2 \\
        \bottomrule
    \end{tabular*}
\end{table}

\subsection{Generalization Across Object Scales and Shapes}\label{sec:exp_gen}

Changes in object geometry can substantially alter contact timing, reachability, and the required whole-body posture. We vary the object scale from $0.4\times$ to $1.6\times$ on all tasks (Push from $0.6\times$ to $1.4\times$), with the nominal training scale at $1.0\times$. For PickUp, we also evaluate four held-out meshes (sphere, stool, cylinder, and torus) and the chair mesh, which appears in training only for other skills and therefore forms a new object--task pairing.

Motion-conditioned baselines often perform well near geometries compatible with their reference motions but degrade when the object scale changes substantially (\cref{tab:unified_sim}). On PickUp, \model maintains strong performance across most scales, with at least 88.0\% success at every scale except $0.4\times$, where its 63.0\% still exceeds the 0.0\% of all three motion-conditioned baselines. On Carry, the \ac{mocap}-based variant achieves the highest scale-averaged success among the compared methods (82.9\% against 81.6\% for PhysHSI), which suggests that the interaction latent is useful for sustained bimanual contact. The trends are more mixed for SitStand and Push, where some motion-conditioned baselines or ablated variants reach higher peak scores at particular scales, such as ResMimic on SitStand and HDMI on Push at $1.0\times$. Reference tracking provides a useful prior where the object geometry is compatible with the reference motion. The Push metric also differs, because it measures sustained hand contact, whereas PickUp and Carry measure completed object manipulation.

The matched ablation in \cref{tab:repr_ablation} isolates the geometric input while keeping the rest of the training pipeline fixed. Raw \ac{df} already outperforms bounding-box and voxel observations overall (81.2\% against 52.5\% and 66.5\%), and encoding the temporal \ac{df} history further raises success to 89.0\%. Bounding-box and voxel success generally increases with object size. Larger objects offer a larger reachable region and tolerate less precise approach behavior, so this trend is consistent with representations that provide useful coarse geometry but encode the contact-relevant link-to-surface relation only indirectly. \Ac{df} observations expose local distance and surface direction directly, which is especially useful for approaching and contacting small objects.

The learned latent is task-structured (\cref{tab:supp_latent_similarity}). At the nominal scale, same-task cosine similarity ranges from 0.39 to 0.55, whereas latents of different tasks are nearly orthogonal; the exception is PickUp and Carry (0.41), which both involve sustained bimanual contact. Same-task similarity is lower across scales (0.08 to 0.28), so the latent is not strictly scale-invariant and preserves scale-dependent changes in interaction.

Shape transfer varies substantially across objects (\cref{tab:supp_ood_repr}). Scale-averaged PickUp success is 55.62\% on the chair and 67.50\% on the stool, against 11.25\%, 10.94\%, and 4.84\% on the sphere, cylinder, and torus, where slippage and rolling before stable contact are common failures. Latent \ac{df} leads on four of the five meshes, whereas raw \ac{df} leads on the chair (61.09\% against 55.62\%). The policy also executes a second new object--task pairing, Push on the chair, in simulation and on hardware.

\subsection{Sequential Multi-Skill Execution}\label{sec:exp_long}

Beyond isolated tasks, we evaluate whether the same deployed policy can execute externally commanded heterogeneous skills without resetting the robot or the environment. Each length-$N$ sequence samples its task instances from the four task types and randomizes the object geometry, pose, and root command of each instance; composed sequences serve only for evaluation (\cref{sec:supp:exp:sequential}). At each skill boundary, the task script updates the interaction-type flag and the root command, and the policy continues from its current physical state. The protocol therefore tests whether one whole-body policy tolerates errors in balance, contact state, object placement, and root tracking that accumulate while it switches between commanded interaction modes. Only variants of the shared policy enter this benchmark, because reference-conditioned methods specify execution through motion and object trajectories and separate task policies would require a different skill-switching mechanism (\cref{sec:supp:exp:baselines}).

The \ac{mocap}-based policy succeeds on 62.1\% of five-task sequences and on 23.5\% of sequences with 15 task instances (\cref{tab:long_horizon}). In a stress test that composes 40 instances of the same four task types under randomized configurations, success falls to 2.1\% but remains nonzero. Success declines quickly with sequence length because a sequence succeeds only if \emph{every} instance succeeds, and individual task success under the randomized distribution is below 100\%. Ablated variants collapse after short sequences, which indicates that the full training stack is important for stable sequential execution.

\begin{figure}[t]
    \centering
    \begin{subfigure}[t]{0.25\linewidth}
        \centering
        \includegraphics[width=\linewidth,height=4cm]{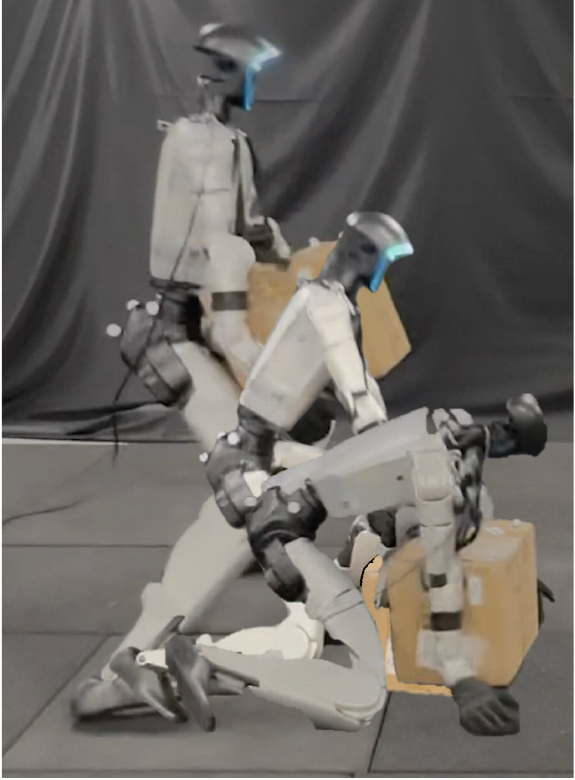}
    \end{subfigure}\hfill
    \begin{subfigure}[t]{0.25\linewidth}
        \centering
        \includegraphics[width=\linewidth,height=4cm]{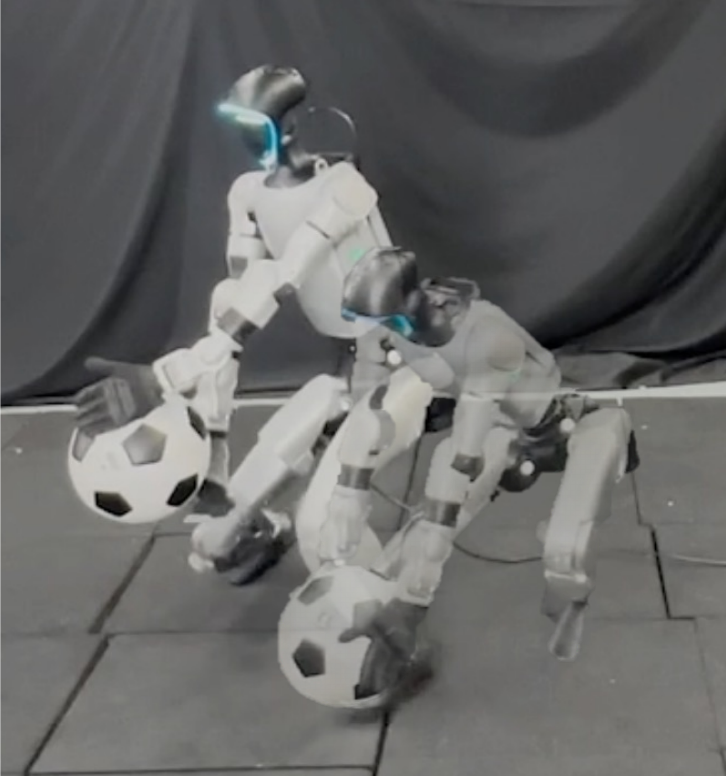}
    \end{subfigure}\hfill
    \begin{subfigure}[t]{0.5\linewidth}
        \centering
        \includegraphics[width=\linewidth,height=4cm]{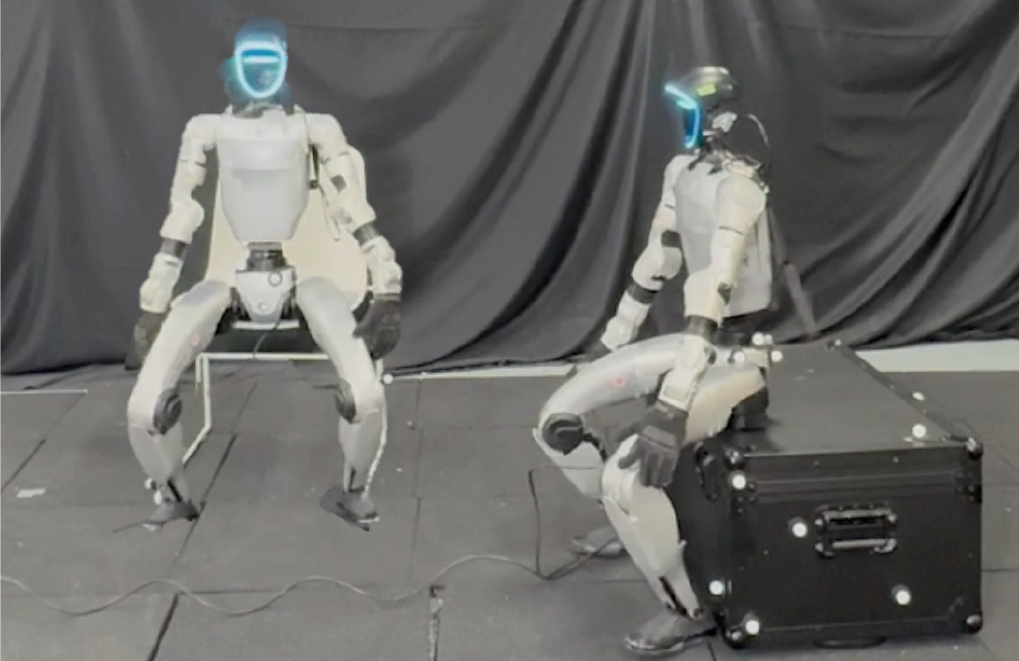}
    \end{subfigure}
    \caption{\textbf{Real-world generalization of \model.} The same policy (a) picks up a box from the training geometry family, (b) picks up a soccer ball, which is an unseen spherical object, and (c) performs SitStand on chairs of two heights ($12\,\mathrm{cm}$ and $46\,\mathrm{cm}$).}
    \label{fig:real}
\end{figure}

\subsection{Visual Distillation Analysis}\label{sec:exp_vis}

The visual encoder is trained on depth observations rendered from three training meshes (a small box, a large case, and a chair) and never on the real-world foam roll or soccer ball used in the held-out deployment tests. The vision policy comes closest to the \ac{mocap}-based policy on PickUp, with success from 63.7\% to 99.7\% across scales (\cref{tab:unified_sim}), but remains substantially weaker on Push, Carry, and sequential execution (\cref{tab:unified_sim,tab:long_horizon}). This gap is expected, because $\pi_{\mathrm{full}}$ receives explicit \ac{df} queries from tracked object geometry, whereas $\pi_{\mathrm{vis}}$ must infer a policy-relevant interaction latent from egocentric depth, which is vulnerable to occlusion and a limited field of view.

\begin{wraptable}{r}{0.48\linewidth}
    \vspace{-8pt}
    \centering
    \small
    \setlength{\tabcolsep}{3pt}
    \caption{\textbf{Sequential multi-skill execution in simulation.} Success rate (\%) over $N$ sequential task instances without resets, reported as mean $\pm$ std over three seeds. Bold marks the best result.}
    \label{tab:long_horizon}
    \resizebox{\linewidth}{!}{%
        \begin{tabular}{lccccc}
            \toprule
            \textbf{Method} & \textbf{N=5} & \textbf{N=10} & \textbf{N=15} & \textbf{N=25} & \textbf{N=40} \\
            \midrule
            \rowcolor{gray!30}
            \multicolumn{6}{l}{\textbf{Ablation study}} \\
            Ours - AIP & \stds{5.2}{0.2} & \stds{0.0}{0.0} & \stds{0.0}{0.0} & \stds{0.0}{0.0} & \stds{0.0}{0.0} \\
            Ours - Syn. & \stds{22.1}{0.8} & \stds{4.9}{0.2} & \stds{1.0}{0.3} & \stds{0.0}{0.0} & \stds{0.0}{0.0} \\
            Ours - Rand. & \stds{1.9}{0.1} & \stds{0.0}{0.0} & \stds{0.0}{0.0} & \stds{0.0}{0.0} & \stds{0.0}{0.0} \\
            Ours - RL & \stds{3.2}{0.2} & \stds{0.0}{0.0} & \stds{0.0}{0.0} & \stds{0.0}{0.0} & \stds{0.0}{0.0} \\
            Ours - Trans. & \stds{1.7}{0.1} & \stds{0.0}{0.0} & \stds{0.0}{0.0} & \stds{0.0}{0.0} & \stds{0.0}{0.0} \\
            \midrule
            \rowcolor{gray!30}
            \multicolumn{6}{l}{\textbf{Our method}} \\
            Ours (MoCap) & \stds{\textbf{62.1}}{1.7} & \stds{\textbf{38.1}}{1.6} & \stds{\textbf{23.5}}{1.2} & \stds{\textbf{9.0}}{0.6} & \stds{\textbf{2.1}}{0.2} \\
            Ours (Vision) & \stds{15.9}{0.6} & \stds{2.5}{0.1} & \stds{0.0}{0.0} & \stds{0.0}{0.0} & \stds{0.0}{0.0} \\
            \bottomrule
        \end{tabular}%
    }%
    \vspace{-10pt}
\end{wraptable}

The standard front camera poorly observes the back-side pelvis--chair contact in SitStand. To assess the role of visual coverage, we retrain the distillation with front and rear depth cameras that jointly cover the interaction. With both cameras, SitStand success reaches 0, 1, 31, 85, and 38\% at $0.4$, $0.6$, $1.0$, $1.4$, and $1.6\times$ (31\% overall), against 0\% at every scale with the front camera alone (\cref{tab:supp_sitstand_fov}). The improvement supports visual coverage as an important factor, and residual partial observability and latent-prediction error may contribute to the remaining gap to the privileged \ac{df} policy. When sequences involving SitStand are excluded, the standard vision policy succeeds on 36\% of sequences, which suggests that the depth-predicted latent can support sequential execution when the relevant contacts are visible (\cref{sec:supp:analysis:vision}).

\subsection{Ablation Study}\label{sec:exp_abla}

Five ablations isolate the role of the major components. \Ac{aip} is particularly important for Carry and sequential execution. Without it (\emph{Ours\,-\,AIP}), the policy fails on Carry and succeeds on only 5.2\% of five-task sequences and on none for $N\ge10$. The same variant obtains higher Push contact rates and higher SitStand success at all three reported scales. With the \ac{df} observation and the rest of the reward stack unchanged, removing the interaction prior lets the remaining rewards favor these contact metrics. The prior thus shapes the balance between individual contact objectives and coordinated multi-phase behavior (\cref{sec:supp:analysis:aip}).

The other ablations identify complementary sources of robustness. Removing synthetic data augmentation (\emph{Ours\,-\,Syn.}) reduces performance on sustained contact and on sequential execution, where five-task success falls from 62.1\% to 22.1\%. These drops indicate that synthetic interaction data extends the coverage of object configurations and contact states beyond the collected demonstrations. Disabling geometry randomization (\emph{Ours\,-\,Rand.}) degrades performance across several tasks and scales, and PickUp fails even at the nominal scale. Removing \ac{rl} fine-tuning (\emph{Ours\,-\,RL}) leaves the behavior-cloned policy with limited robustness under geometric variation. Replacing the Transformer with a \ac{mlp} backbone (\emph{Ours\,-\,Trans.}) degrades the tasks that require temporal coordination. This variant fails PickUp at every scale, fails Carry, and succeeds on only 1.7\% of five-task sequences.

\subsection{Real-World Deployment}\label{sec:exp_real}

\begin{wrapfigure}{r}{0.48\linewidth}
    \vspace{-8pt}
    \centering
    \includegraphics[width=\linewidth]{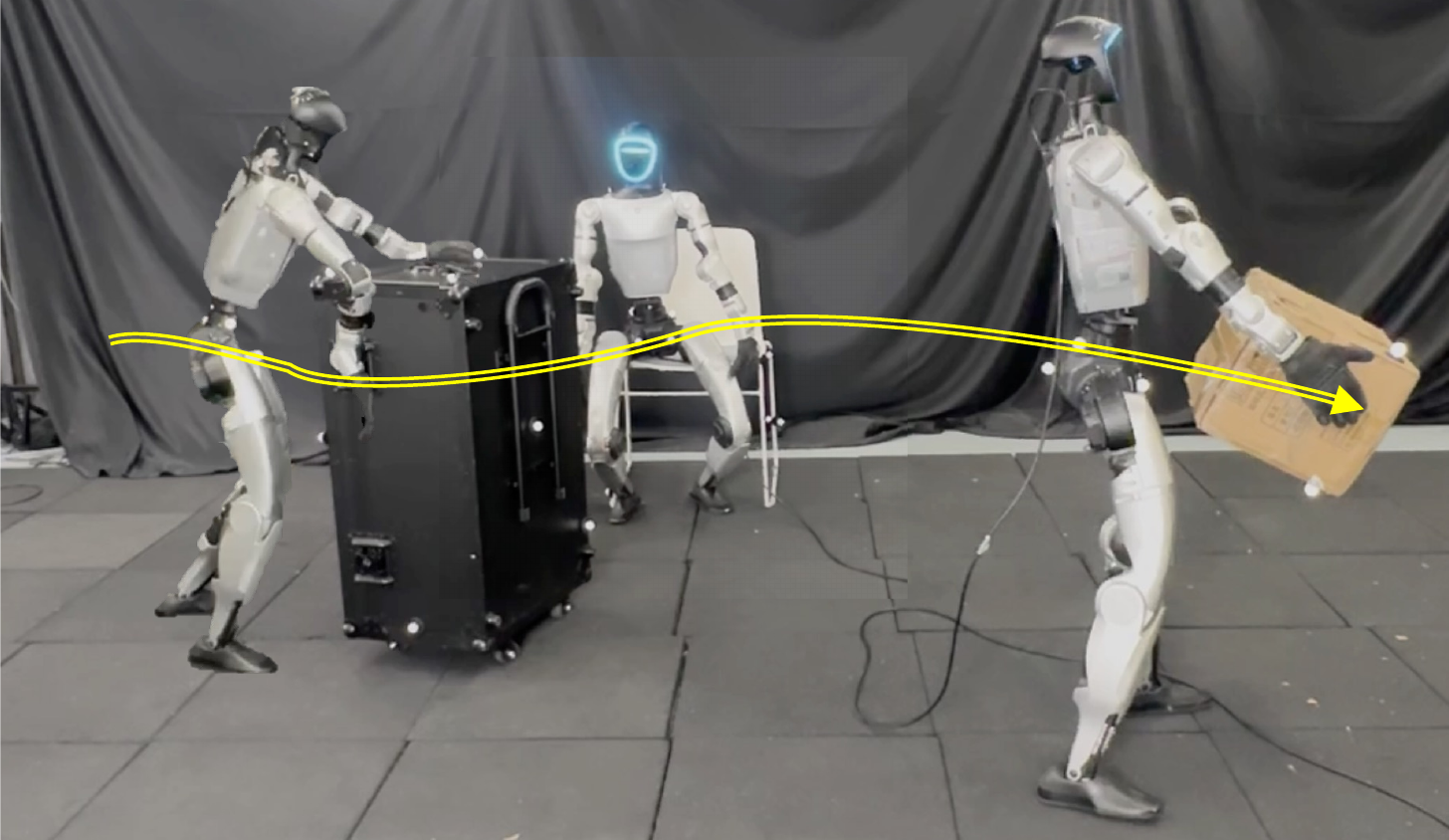}
    \caption{\textbf{Qualitative real-world sequential execution.} A single \model policy performs Push, PickUp, and Carry in sequence without resets, following externally provided task and root commands.}
    \label{fig:longterm}
    \vspace{-10pt}
\end{wrapfigure}

We evaluate \model on a Unitree G1 with qualitative demonstrations and quantitative repeated trials. Qualitative tests cover all four skills and sequential execution (\cref{fig:real,fig:longterm}) and provide evidence that the same deployed policy interface transfers across the full task suite on hardware. Quantitative trials cover the two tasks with reliable and repeatable measurement protocols, PickUp across two object sizes and SitStand across two chair heights (\cref{sec:supp:exp:real}). External \ac{mocap}, used for evaluation only, measures the root-tracking accuracy $R_{\mathrm{acc}}$, the fraction of frames with planar root-position error below $5\,\mathrm{cm}$.

The \ac{mocap}-based policy transfers reliably to hardware in these settings, succeeding in 10 and 8 of 10 PickUp trials and in 8 and 10 of 10 SitStand trials (\cref{tab:deploy}). The vision policy succeeds in both PickUp settings, in 8 and 7 of 10 trials, with lower success and tracking accuracy, consistent with the simulation gap from partial egocentric observability. It is not evaluated on SitStand because its camera poorly observes the back-side contact.

\section{Conclusion}

In \model, grounding control in local robot--object geometry rather than in deployed motion references lets one humanoid policy adapt to object-scale variation and new interaction settings, follow flexible commands, and compose heterogeneous skills in sequence. Unified \ac{df} representations and a three-stage pipeline of teacher-guided behavior cloning, \ac{aip}-guided \ac{rl}, and visual-motor distillation turn this idea into policies that are motion-free at inference and deployable with either \ac{mocap}-based geometry or egocentric depth. In simulation, \model generalizes across four tasks and object scales from $0.4\times$ to $1.6\times$, succeeds on 62.1\% of five-task sequences, and retains 23.5\% success for 15 task instances. Hardware experiments show transfer across object geometries, with quantitative trials on PickUp and SitStand and qualitative demonstrations of the full task suite.

Three limitations remain (\cref{sec:supp:exp:failure}). First, the per-link \ac{df} captures coarse proximity and surface direction rather than a detailed pressure distribution, so tasks that need precise force balance or fine contact placement remain difficult; force-aware feedback and denser geometric queries could help. Second, vision-based failures occur mainly when task-relevant contacts are occluded or outside the egocentric view, as for the back-side chair contact in SitStand; rear-facing or multi-view sensing could cover such contacts. Third, sensing latency can delay contact during fast motions, and the Python deployment stack adds roughly 5 to 15\,ms of variable control latency; lower-latency perception and an optimized C++/ONNX control stack could reduce both.

\begin{ack}
This work is supported in part by the Brain Science and Brain-like Intelligence Technology---National Science and Technology Major Project (2025ZD0219400), the National Natural Science Foundation of China (62376009), the Beijing Nova program, the NVIDIA Academic Grant Program using Spark and Thor, the PKU-BingJi Joint Laboratory for Artificial Intelligence, the Wuhan Major Scientific and Technological Special Program (2025060902020304), the Hubei Embodied Intelligence Foundation Model Research and Development Program, and the National Comprehensive Experimental Base for Governance of Intelligent Society, Wuhan East Lake High-Tech Development Zone. The authors declare no competing interests.
\end{ack}

\FloatBarrier
{\small
\bibliographystyle{apalike}
\bibliography{reference_header,references}}

@string {NIPS = "{Proceedings of Advances in Neural Information Processing Systems (NeurIPS)}"}

@string {NIPSBENCH = "{Proceedings of the Neural Information Processing Systems (NeurIPS) Track on Datasets and Benchmarks}"}

@string {ICLR = "{Proceedings of International Conference on Learning Representations (ICLR)}"}

@string {AISTATS = "{International Conference on Artificial Intelligence and Statistics (AISTATS)}"}

@string {IJRR = "{International Journal of Robotics Research (IJRR)}"}

@string {RA-L = "{IEEE Robotics and Automation Letters (RA-L)}"}

@string {IROS = "{IEEE/RSJ International Conference on Intelligent Robots and Systems (IROS)}"}

@string {ICRA = "{IEEE International Conference on Robotics and Automation (ICRA)}"}

@string {RSS = "{Robotics: Science and Systems (RSS)}"}

@string {CoRL = "{Conference on Robot Learning (CoRL)}"}

@string {TOG = "{ACM Transactions on Graphics (TOG)}"}

@string {SIGGRAPH = "ACM SIGGRAPH Conference Proceedings"}

@string {SIGGRAPHASIA = "ACM SIGGRAPH Asia Conference Proceedings"}

@string {CHI = "{ACM Conference on Human Factors in Computing Systems (CHI)}"}

@inproceedings{he2024omnih2o,
  title={{OmniH2O}: Universal and dexterous human-to-humanoid whole-body teleoperation and learning},
  author={He, Tairan and Luo, Zhengyi and He, Xialin and Xiao, Wenli and Zhang, Chong and Zhang, Weinan and Kitani, Kris and Liu, Changliu and Shi, Guanya},
  booktitle=CoRL,
  year={2024}
}

@inproceedings{li2025clone,
  title={{CLONE}: Closed-Loop Whole-Body Humanoid Teleoperation for Long-Horizon Tasks},
  author={Li, Yixuan and Lin, Yutang and Cui, Jieming and Liu, Tengyu and Liang, Wei and Zhu, Yixin and Huang, Siyuan},
  booktitle=CoRL,
  year={2025}
}

@inproceedings{tessler2023calm,
  title={{CALM}: Conditional adversarial latent models for directable virtual characters},
  author={Tessler, Chen and Kasten, Yoni and Guo, Yunrong and Mannor, Shie and Chechik, Gal and Peng, Xue Bin},
  booktitle=SIGGRAPH,
  year={2023}
}

@inproceedings{ross2011reduction,
  title={A reduction of imitation learning and structured prediction to no-regret online learning},
  author={Ross, St{\'e}phane and Gordon, Geoffrey and Bagnell, Drew},
  booktitle=AISTATS,
  year={2011}
}

@article{peng2021amp,
  title={{AMP}: Adversarial motion priors for stylized physics-based character control},
  author={Peng, Xue Bin and Ma, Ze and Abbeel, Pieter and Levine, Sergey and Kanazawa, Angjoo},
  journal=TOG,
  volume={40},
  number={4},
  pages={1--20},
  year={2021},
  publisher={ACM}
}

@article{liao2025beyond,
  title={{BeyondMimic}: From Motion Tracking to Versatile Humanoid Control via Guided Diffusion},
  author={Qiayuan Liao and Takara E. Truong and Xiaoyu Huang and Yuman Gao and Guy Tevet and Koushil Sreenath and C. Karen Liu},
  journal={Science Robotics},
  volume={11},
  number={117},
  pages={eadx8924},
  year={2026},
  doi={10.1126/scirobotics.adx8924}
}

@inproceedings{zhang2025track,
  title={Track Any Motions under Any Disturbances},
  author={Zhikai Zhang and Jun Guo and Chao Chen and Jilong Wang and Chenghuai Lin and Yunrui Lian and Han Xue and Zhenrong Wang and Maoqi Liu and Jiangran Lyu and Huaping Liu and He Wang and Li Yi},
  booktitle=ICRA,
  year={2026},
  pages={17533--17540}
}

@inproceedings{li2025bfm,
  title={{BFM-Zero}: A Promptable Behavioral Foundation Model for Humanoid Control Using Unsupervised Reinforcement Learning},
  author={Yitang Li and Zhengyi Luo and Tonghe Zhang and Cunxi Dai and Anssi Kanervisto and Andrea Tirinzoni and Haoyang Weng and Kris Kitani and Mateusz Guzek and Ahmed Touati and Alessandro Lazaric and Matteo Pirotta and Guanya Shi},
  booktitle=ICLR,
  year={2026}
}

@inproceedings{du2025learning,
  title={{COLA}: Learning Human-Humanoid Coordination for Collaborative Object Carrying},
  author={Yushi Du and Yixuan Li and Baoxiong Jia and Yutang Lin and Pei Zhou and Wei Liang and Yanchao Yang and Siyuan Huang},
  booktitle=ICRA,
  year={2026}
}

@inproceedings{yang2025omniretarget,
  title={{OmniRetarget}: Interaction-Preserving Data Generation for Humanoid Whole-Body Loco-Manipulation and Scene Interaction},
  author={Lujie Yang and Xiaoyu Huang and Zhen Wu and Angjoo Kanazawa and Pieter Abbeel and Carmelo Sferrazza and C. Karen Liu and Rocky Duan and Guanya Shi},
  booktitle=ICRA,
  year={2026},
  doi={10.1109/ICRA57385.2026.11696487},
  pages={11691--11698}
}

@article{weng2025hdmi,
  title={{HDMI}: Learning Interactive Humanoid Whole-Body Control from Human Videos},
  author={Haoyang Weng and Yitang Li and Nikhil Sobanbabu and Zihan Wang and Zhengyi Luo and Tairan He and Deva Ramanan and Guanya Shi},
  journal={arXiv preprint arXiv:2509.16757},
  year={2025}
}

@article{xue2025leverb,
  title={{LeVERB}: Humanoid Whole-Body Control with Latent Vision-Language Instruction},
  author={Haoru Xue and Xiaoyu Huang and Dantong Niu and Qiayuan Liao and Thomas Kragerud and Jan Tommy Gravdahl and Xue Bin Peng and Guanya Shi and Trevor Darrell and Koushil Sreenath and Shankar Sastry},
  journal={arXiv preprint arXiv:2506.13751},
  year={2025}
}

@article{luo2025sonic,
  title={{SONIC}: Supersizing Motion Tracking for Natural Humanoid Whole-Body Control},
  author={Luo, Zhengyi and Yuan, Ye and Wang, Tingwu and Li, Chenran and Casta{\~n}eda, Fernando and Chen, Sirui and Cao, Zi-Ang and Li, Jiefeng and Minor, David and Ben, Qingwei and Park, Jinhyung and Sami, David and Wang, Zi and Da, Xingye and Ding, Runyu and Hogg, Cyrus and Song, Lina and Lim, Edy and Jeong, Eugene and He, Tairan and Xue, Haoru and Xiao, Wenli and Yuen, Simon and Kautz, Jan and Chang, Yan and Iqbal, Umar and Fan, Linxi and Zhu, Yuke},
  journal={Science Robotics},
  volume={11},
  number={117},
  pages={eaed4592},
  year={2026},
  doi={10.1126/scirobotics.aed4592}
}

@inproceedings{ze2025twist2,
  title={{TWIST2}: Scalable, Portable, and Holistic Humanoid Data Collection System},
  author={Yanjie Ze and Siheng Zhao and Weizhuo Wang and Angjoo Kanazawa and Rocky Duan and Pieter Abbeel and Guanya Shi and Jiajun Wu and C. Karen Liu},
  booktitle=ICRA,
  year={2026}
}

@inproceedings{zhao2025resmimic,
  title={{ResMimic}: From General Motion Tracking to Humanoid Whole-body Loco-Manipulation via Residual Learning},
  author={Siheng Zhao and Yanjie Ze and Yue Wang and C. Karen Liu and Pieter Abbeel and Guanya Shi and Rocky Duan},
  booktitle={{IEEE-RAS International Conference on Humanoid Robots (Humanoids)}},
  year={2026}
}

@article{yin2025visual,
  title={{VisualMimic}: Visual Humanoid Loco-Manipulation via Motion Tracking and Generation},
  author={Shaofeng Yin and Yanjie Ze and Hong-Xing Yu and C. Karen Liu and Jiajun Wu},
  journal={arXiv preprint arXiv:2509.20322},
  year={2025}
}

@article{fu2025demohlm,
  author={Yuhui Fu and Feiyang Xie and Chaoyi Xu and Jing Xiong and Haoqi Yuan and Zongqing Lu},
  journal=RA-L,
  title={{DemoHLM}: From One Demonstration to Generalizable Humanoid Loco-Manipulation},
  year={2026},
  volume={11},
  number={4},
  pages={4393--4400},
  doi={10.1109/LRA.2026.3666395}
}

@article{wang2025physhsi,
  title={{PhysHSI}: Towards a Real-World Generalizable and Natural Humanoid-Scene Interaction System},
  author={Huayi Wang and Wentao Zhang and Runyi Yu and Tao Huang and Junli Ren and Feiyu Jia and Zirui Wang and Xiaojie Niu and Xiao Chen and Jiahe Chen and Qifeng Chen and Jingbo Wang and Jiangmiao Pang},
  journal={arXiv preprint arXiv:2510.11072},
  year={2025}
}

@article{nakamura1987task,
  title={Task-priority based redundancy control of robot manipulators},
  author={Nakamura, Yoshihiko and Hanafusa, Hideo and Yoshikawa, Tsuneo},
  journal=IJRR,
  volume={6},
  number={2},
  pages={3--15},
  year={1987},
  publisher={SAGE Publications}
}

@article{khatib2004whole,
  title={Whole-body dynamic behavior and control of human-like robots},
  author={Khatib, Oussama and Sentis, Luis and Park, Jaeheung and Warren, James},
  journal={International Journal of Humanoid Robotics},
  volume={1},
  number={01},
  pages={29--43},
  year={2004},
  publisher={World Scientific}
}

@article{dietrich2015overview,
  title={An overview of null space projections for redundant, torque-controlled robots},
  author={Dietrich, Alexander and Ott, Christian and Albu-Sch{\"a}ffer, Alin},
  journal=IJRR,
  volume={34},
  number={11},
  pages={1385--1400},
  year={2015},
  publisher={SAGE Publications}
}

@incollection{moro2019whole,
  title={Whole-body control of humanoid robots},
  author={Moro, Federico L and Sentis, Luis},
  booktitle={Humanoid Robotics: a Reference},
  pages={1161--1183},
  year={2019},
  publisher={Springer Dordrecht}
}

@article{sentis2005synthesis,
  title={Synthesis of whole-body behaviors through hierarchical control of behavioral primitives},
  author={Sentis, Luis and Khatib, Oussama},
  journal={International Journal of Humanoid Robotics},
  volume={2},
  number={04},
  pages={505--518},
  year={2005},
  publisher={World Scientific}
}

@inproceedings{fu2024humanplus,
  author={Fu, Zipeng and Zhao, Qingqing and Wu, Qi and Wetzstein, Gordon and Finn, Chelsea},
  title={{HumanPlus}: Humanoid Shadowing and Imitation from Humans},
  booktitle=CoRL,
  year={2024}
}

@inproceedings{ji2024exbody2,
  title={{ExBody2}: Advanced expressive humanoid whole-body control},
  author={Ji, Mazeyu and Peng, Xuanbin and Liu, Fangchen and Li, Jialong and Yang, Ge and Cheng, Xuxin and Wang, Xiaolong},
  booktitle=ICRA,
  year={2026},
  doi={10.1109/ICRA57385.2026.11696204}
}

@inproceedings{allshire2025visual,
  title={Visual Imitation Enables Contextual Humanoid Control},
  author={Allshire, Arthur and Choi, Hongsuk and Zhang, Junyi and McAllister, David and Zhang, Anthony and Kim, Chung Min and Darrell, Trevor and Abbeel, Pieter and Malik, Jitendra and Kanazawa, Angjoo},
  booktitle=CoRL,
  year={2025}
}

@inproceedings{ze2025twist,
  title={{TWIST}: Teleoperated whole-body imitation system},
  author={Ze, Yanjie and Chen, Zixuan and Ara{\'u}jo, Joao Pedro and Cao, Zi-ang and Peng, Xue Bin and Wu, Jiajun and Liu, C Karen},
  booktitle=CoRL,
  year={2025}
}

@inproceedings{li2025amo,
  title={{AMO}: Adaptive Motion Optimization for Hyper-Dexterous Humanoid Whole-Body Control},
  author={Li, Jialong and Cheng, Xuxin and Huang, Tianshu and Yang, Shiqi and Qiu, Ri-Zhao and Wang, Xiaolong},
  booktitle=RSS,
  year={2025}
}

@inproceedings{he2024learning,
  title={Learning human-to-humanoid real-time whole-body teleoperation},
  author={He, Tairan and Luo, Zhengyi and Xiao, Wenli and Zhang, Chong and Kitani, Kris and Liu, Changliu and Shi, Guanya},
  booktitle=IROS,
  year={2024}
}

@article{lunardi2023reference,
  title={Reference-Free Model Predictive Control for Quadrupedal Locomotion},
  author={Lunardi, Gianni and Corb{\`e}res, Thomas and Mastalli, Carlos and Mansard, Nicolas and Flayols, Thomas and Tonneau, Steve and Del Prete, Andrea},
  journal={IEEE Access},
  volume={12},
  pages={689--698},
  year={2024},
  publisher={IEEE},
  doi={10.1109/ACCESS.2023.3345157}
}

@article{osher1988fronts,
  title={Fronts propagating with curvature-dependent speed: Algorithms based on {Hamilton-Jacobi} formulations},
  author={Osher, Stanley and Sethian, James A},
  journal={Journal of Computational Physics},
  volume={79},
  number={1},
  pages={12--49},
  year={1988},
  publisher={Elsevier}
}

@inproceedings{posa2013direct,
  title={Direct trajectory optimization of rigid body dynamical systems through contact},
  author={Posa, Michael and Tedrake, Russ},
  booktitle={Tenth Workshop on the Algorithmic Foundations of Robotics},
  year={2013}
}

@article{schulman2014motion,
  title={Motion planning with sequential convex optimization and convex collision checking},
  author={Schulman, John and Duan, Yan and Ho, Jonathan and Lee, Alex and Awwal, Ibrahim and Bradlow, Henry and Pan, Jia and Patil, Sachin and Goldberg, Ken and Abbeel, Pieter},
  journal=IJRR,
  volume={33},
  number={9},
  pages={1251--1270},
  year={2014},
  publisher={SAGE Publications}
}

@article{herzog2016momentum,
  title={Momentum control with hierarchical inverse dynamics on a torque-controlled humanoid},
  author={Herzog, Alexander and Rotella, Nicholas and Mason, Sean and Grimminger, Felix and Schaal, Stefan and Righetti, Ludovic},
  journal={Autonomous Robots},
  volume={40},
  number={3},
  pages={473--491},
  year={2016},
  publisher={Springer}
}

@article{hopkins2016optimization,
  title={Optimization-based whole-body control of a series elastic humanoid robot},
  author={Hopkins, Michael A and Leonessa, Alexander and Lattimer, Brian Y and Hong, Dennis W},
  journal={International Journal of Humanoid Robotics},
  volume={13},
  number={01},
  pages={1550034},
  year={2016},
  publisher={World Scientific}
}

@inproceedings{zhang2023efficient,
  title={Efficient sim-to-real transfer of contact-rich manipulation skills with online admittance residual learning},
  author={Zhang, Xiang and Wang, Changhao and Sun, Lingfeng and Wu, Zheng and Zhu, Xinghao and Tomizuka, Masayoshi},
  booktitle=CoRL,
  year={2023}
}

@inproceedings{kumar2021rma,
  title={{RMA}: Rapid motor adaptation for legged robots},
  author={Kumar, Ashish and Fu, Zipeng and Pathak, Deepak and Malik, Jitendra},
  booktitle=RSS,
  year={2021}
}

@inproceedings{li2025hold,
  title={Hold My Beer: Learning Gentle Humanoid Locomotion and End-Effector Stabilization Control},
  author={Li, Yitang and Zhang, Yuanhang and Xiao, Wenli and Pan, Chaoyi and Weng, Haoyang and He, Guanqi and He, Tairan and Shi, Guanya},
  booktitle=CoRL,
  year={2025}
}

@article{haarnoja2024learning,
  title={Learning agile soccer skills for a bipedal robot with deep reinforcement learning},
  author={Haarnoja, Tuomas and Moran, Ben and Lever, Guy and Huang, Sandy H. and Tirumala, Dhruva and Humplik, Jan and Wulfmeier, Markus and Tunyasuvunakool, Saran and Siegel, Noah Y. and Hafner, Roland and Bloesch, Michael and Hartikainen, Kristian and Byravan, Arunkumar and Hasenclever, Leonard and Tassa, Yuval and Sadeghi, Fereshteh and Batchelor, Nathan and Casarini, Federico and Saliceti, Stefano and Game, Charles and Sreendra, Neil and Patel, Kushal and Gwira, Marlon and Huber, Andrea and Hurley, Nicole and Nori, Francesco and Hadsell, Raia and Heess, Nicolas},
  journal={Science Robotics},
  volume={9},
  number={89},
  pages={eadi8022},
  year={2024},
  publisher={American Association for the Advancement of Science},
  doi={10.1126/scirobotics.adi8022}
}

@inproceedings{huang2023voxposer,
  title={{VoxPoser}: Composable {3D} value maps for robotic manipulation with language models},
  author={Huang, Wenlong and Wang, Chen and Zhang, Ruohan and Li, Yunzhu and Wu, Jiajun and Fei-Fei, Li},
  booktitle=CoRL,
  year={2023}
}

@inproceedings{xu2020learning,
  title={Learning {3D} dynamic scene representations for robot manipulation},
  author={Xu, Zhenjia and He, Zhanpeng and Wu, Jiajun and Song, Shuran},
  booktitle=CoRL,
  year={2020}
}

@inproceedings{varley2017shape,
  title={Shape completion enabled robotic grasping},
  author={Varley, Jacob and DeChant, Chad and Richardson, Adam and Ruales, Joaqu{\'\i}n and Allen, Peter},
  booktitle=IROS,
  year={2017}
}

@inproceedings{jiang2021synergies,
  title={Synergies between affordance and geometry: {6-DoF} grasp detection via implicit representations},
  author={Jiang, Zhenyu and Zhu, Yifeng and Svetlik, Maxwell and Fang, Kuan and Zhu, Yuke},
  booktitle=RSS,
  year={2021}
}

@inproceedings{liu2025fetchbot,
  title={{FetchBot}: Learning generalizable object fetching in cluttered scenes via zero-shot {Sim2Real}},
  author={Liu, Weiheng and Wan, Yuxuan and Wang, Jilong and Kuang, Yuxuan and Shi, Xuesong and Li, Haoran and Zhao, Dongbin and Zhang, Zhizheng and Wang, He},
  booktitle=CoRL,
  year={2025}
}

@inproceedings{liu2022frame,
  title={Frame mining: a free lunch for learning robotic manipulation from {3D} point clouds},
  author={Liu, Minghua and Li, Xuanlin and Ling, Zhan and Li, Yangyan and Su, Hao},
  booktitle=CoRL,
  year={2022}
}

@inproceedings{zhu2024point,
  title={Point cloud matters: Rethinking the impact of different observation spaces on robot learning},
  author={Zhu, Haoyi and Wang, Yating and Huang, Di and Ye, Weicai and Ouyang, Wanli and He, Tong},
  booktitle=NIPSBENCH,
  year={2024},
  doi={10.52202/079017-2473}
}

@inproceedings{chisari2024learning,
  title={Learning robotic manipulation policies from point clouds with conditional flow matching},
  author={Chisari, Eugenio and Heppert, Nick and Argus, Max and Welschehold, Tim and Brox, Thomas and Valada, Abhinav},
  booktitle=CoRL,
  year={2024}
}

@inproceedings{jia2025pointmappolicy,
  title={{PointMapPolicy}: Structured Point Cloud Processing for Multi-Modal Imitation Learning},
  author={Jia, Xiaogang and Wang, Qian and Wang, Anrui and Wang, Han A. and Gyenes, Bal{\'a}zs and Gospodinov, Emiliyan and Jiang, Xinkai and Li, Ge and Zhou, Hongyi and Liao, Weiran and Huang, Xi and Beck, Maximilian and Reuss, Moritz and Lioutikov, Rudolf and Neumann, Gerhard},
  booktitle=NIPS,
  year={2025},
  doi={10.52202/085713-5353}
}

@inproceedings{shen2021igibson,
  title={{iGibson} 1.0: A simulation environment for interactive tasks in large realistic scenes},
  author={Shen, Bokui and Xia, Fei and Li, Chengshu and Mart{\'i}n-Mart{\'i}n, Roberto and Fan, Linxi and Wang, Guanzhi and P{\'e}rez-D'Arpino, Claudia and Buch, Shyamal and Srivastava, Sanjana and Tchapmi, Lyne P. and Tchapmi, Micael E. and Vainio, Kent and Wong, Josiah and Li, Fei-Fei and Savarese, Silvio},
  booktitle=IROS,
  year={2021},
  doi={10.1109/IROS51168.2021.9636667},
  pages={7520--7527}
}

@inproceedings{pan2012fcl,
  title={{FCL}: A general purpose library for collision and proximity queries},
  author={Pan, Jia and Chitta, Sachin and Manocha, Dinesh},
  booktitle=ICRA,
  year={2012}
}

@inproceedings{hauser2016robust,
  title={Robust contact generation for robot simulation with unstructured meshes},
  author={Hauser, Kris},
  booktitle={International Symposium on Robotics Research},
  year={2016}
}

@inproceedings{zesch2023neural,
  title={Neural collision fields for triangle primitives},
  author={Zesch, Ryan S and Modi, Vismay and Sueda, Shinjiro and Levin, David IW},
  booktitle=SIGGRAPHASIA,
  year={2023}
}

@inproceedings{dai2022graspnerf,
  title={{GraspNeRF}: Multiview-based {6-DoF} grasp detection for transparent and specular objects using generalizable {NeRF}},
  author={Dai, Qiyu and Zhu, Yan and Geng, Yiran and Ruan, Ciyu and Zhang, Jiazhao and Wang, He},
  booktitle=ICRA,
  year={2023}
}

@inproceedings{quintero2024stochastic,
  title={Stochastic implicit neural signed distance functions for safe motion planning under sensing uncertainty},
  author={Quintero-Pena, Carlos and Thomason, Wil and Kingston, Zachary and Kyrillidis, Anastasios and Kavraki, Lydia E},
  booktitle=ICRA,
  year={2024}
}

@inproceedings{song2025implicit,
  title={Implicit Neural-Representation Learning for Elastic Deformable-Object Manipulations},
  author={Song, Minseok and Ha, JeongHo and Park, Bonggyeong and Park, Daehyung},
  booktitle=RSS,
  year={2025},
  doi={10.15607/RSS.2025.XXI.035}
}

@inproceedings{li2024rdf,
  title={Representing Robot Geometry as Distance Fields: Applications to Whole-body Manipulation},
  author={Li, Yiming and Zhang, Yan and Razmjoo, Amirreza and Calinon, Sylvain},
  booktitle=ICRA,
  year={2024}
}

@article{yang2025contactsdf,
  title={{ContactSDF}: Signed distance functions as multi-contact models for dexterous manipulation},
  author={Yang, Wen and Jin, Wanxin},
  journal=RA-L,
  volume={10},
  number={5},
  pages={4212--4219},
  year={2025},
  publisher={IEEE},
  doi={10.1109/LRA.2025.3548503}
}

@inproceedings{wang2026humanx,
  title={{HumanX}: Toward Agile and Generalizable Humanoid Interaction Skills from Human Videos},
  author={Wang, Yinhuai and Zhao, Qihan and Lau, Yuen Fui and Yu, Runyi and Tsui, Hok Wai and Chen, Qifeng and Wang, Jingbo and Pang, Jiangmiao and Tan, Ping},
  booktitle=CoRL,
  year={2026}
}

@inproceedings{li2026haic,
  title={{HAIC}: Humanoid Agile Object Interaction Control via Dynamics-Aware World Model},
  author={Li, Dongting and Chen, Xingyu and Wu, Qianyang and Chen, Bo and Wu, Sikai and Wu, Hanyu and Zhang, Guoyao and Li, Liang and Zhou, Mingliang and Xiang, Diyun and Ma, Jianzhu and Zhang, Qiang and Xu, Renjing},
  booktitle=RSS,
  year={2026},
  doi={10.15607/RSS.2026.XXII.013}
}

@inproceedings{zheng2025bulky,
  title={Embracing Bulky Objects with Humanoid Robots: Whole-Body Manipulation with Reinforcement Learning},
  author={Zheng, Chunxin and Chen, Kai and Bi, Zhihai and Li, Yulin and Pan, Liang and Zhou, Jinni and Li, Haoang and Ma, Jun},
  booktitle=ICRA,
  year={2026},
  pages={6089--6096}
}

@inproceedings{wang2025experts,
  title={From experts to a generalist: Toward general whole-body control for humanoid robots},
  author={Wang, Yuxuan and Yang, Ming and Ding, Ziluo and Zhang, Yu and Zeng, Weishuai and Xu, Xinrun and Jiang, Haobin and Lu, Zongqing},
  booktitle=NIPS,
  year={2025}
}

@inproceedings{xie2025kungfubot,
  title={{KungfuBot}: Physics-based humanoid whole-body control for learning highly-dynamic skills},
  author={Xie, Weiji and Han, Jinrui and Zheng, Jiakun and Li, Huanyu and Liu, Xinzhe and Shi, Jiyuan and Zhang, Weinan and Bai, Chenjia and Li, Xuelong},
  booktitle=NIPS,
  year={2025}
}

@inproceedings{shi2025adversarial,
  title={Adversarial locomotion and motion imitation for humanoid policy learning},
  author={Shi, Jiyuan and Liu, Xinzhe and Wang, Dewei and Lu, Ouyang and Schwertfeger, S{\"o}ren and Zhang, Chi and Sun, Fuchun and Bai, Chenjia and Li, Xuelong},
  booktitle=NIPS,
  year={2025}
}

@inproceedings{jing2025humanoidgen,
  title={{HumanoidGen}: Data generation for bimanual dexterous manipulation via {LLM} reasoning},
  author={Jing, Zhi and Yang, Siyuan and Ao, Jicong and Xiao, Ting and Jiang, Yu-Gang and Bai, Chenjia},
  booktitle=NIPS,
  year={2025}
}

\clearpage
\acresetall
\appendix
\renewcommand\thefigure{A\arabic{figure}}
\setcounter{figure}{0}
\renewcommand\thetable{A\arabic{table}}
\setcounter{table}{0}
\renewcommand\theequation{A\arabic{equation}}
\setcounter{equation}{0}
\pagenumbering{arabic}
\renewcommand*{\thepage}{A\arabic{page}}
\setcounter{footnote}{0}

\section{Implementation Details}\label{sec:supp:method}

This section details the interaction representation, policy architecture, training stages, reward formulation, and visual-motor distillation of the \model pipeline introduced in \cref{sec:method}. Unless stated otherwise, all policy variants share the same whole-body control architecture and differ only in the sensing input or the training objective.

\subsection{Interaction Representation and Encoder}\label{sec:supp:method:interaction}

We compute distance-field queries from the object geometry with NVIDIA Kaolin. In the real-world setup with \ac{mocap}, the object mesh and rigid pose are known, so no online mesh reconstruction is required; the queried humanoid links are transformed into the object frame before a batched point-to-surface evaluation. On an RTX 5090, the measured latency stays below $9\,\mathrm{ms}$ per pass even for 1024 parallel environments with three objects (\cref{tab:supp_df_latency}). Queries are evaluated at a fixed set of humanoid links shared across all skills, namely both wrists, both ankles, and the pelvis. The fixed query set avoids task-specific observation layouts and keeps the policy interface identical across PickUp, SitStand, Push, and Carry.

\begin{table}[b!]
    \centering
    \small
    \caption{\textbf{Latency of distance-field queries on one RTX 5090.} Time per pass (ms) for 1, 7, and 30 queries.}
    \label{tab:supp_df_latency}
    \begin{tabular*}{\linewidth}{@{\extracolsep{\fill}}llrrr@{}}
        \toprule
        \textbf{Environments} & \textbf{Objects} & \textbf{1 query} & \textbf{7 queries} & \textbf{30 queries} \\
        \midrule
        1 & 1 & 3.449 & 3.260 & 3.313 \\
        1 & 2 & 6.135 & 5.832 & 6.002 \\
        1 & 3 & 8.784 & 8.898 & 8.712 \\
        \midrule
        1024 & 1 & 3.230 & 3.283 & 3.278 \\
        1024 & 2 & 6.000 & 6.007 & 6.043 \\
        1024 & 3 & 8.720 & 8.808 & 8.776 \\
        \bottomrule
    \end{tabular*}
\end{table}

For each queried link, the deployable geometric observation contains the distance to the nearest object surface and a local surface-direction estimate. At differentiable points of the distance field, the direction is the normalized gradient. At non-smooth regions, such as sharp edges, corners, or medial-axis-like regions where the nearest surface projection is not unique, we use a normalized numerical direction estimate instead. When this estimate changes abruptly, we retain the last valid direction and rely on the encoder of the \ac{vae} to smooth the resulting local discontinuity before the policy consumes the feature.

The interaction history spans a temporal window of length $l=5$ and is encoded into a 64-dimensional latent $z_t$. Although the raw feature is already compact, the \ac{vae} does more than reduce dimensionality. It smooths discontinuous local geometry cues, regularizes the interaction representation, and provides a stable latent space for policy conditioning and adversarial interaction regularization.

The encoder predicts a Gaussian posterior
\begin{equation}
    q_\phi(z_t\mid I_t)=\mathcal{N}(\mu_\phi(I_t),\sigma_\phi(I_t)),
\end{equation}
the decoder reconstructs the interaction history, and the \ac{vae} is trained with
\begin{equation}
    \mathcal{L}_{\mathrm{VAE}}=\|\hat I_t-I_t\|_2^2+\beta D_{\mathrm{KL}}\left(q_\phi(z_t\mid I_t)\,\|\,\mathcal{N}(0,I)\right).
\end{equation}
The \ac{vae} is trained jointly with the policy during interaction skill pre-training under the DAgger behavior-cloning objective, so gradients from both the reconstruction objective and policy learning update the interaction encoder. After pre-training, the encoder is frozen and produces $z_t$ during discriminative post-training. During visual-motor distillation, the frozen \ac{df} encoder defines the teacher-side latent, while the visual encoder learns to predict a policy-relevant replacement latent from egocentric depth.

During discriminative post-training with \ac{aip}, we augment the \ac{df} latent with link velocities decomposed in the local surface frame into a normal component along the local \ac{df} direction and a tangential remainder (\cref{sec:method:posttrain}). The deployed policy thus remains conditioned on $z_t$ without these velocities, whereas the \ac{aip} discriminator receives the velocity-augmented statistic $h_t=[z_t,\{\mathbf{v}^{\mathrm{norm}}_{t,k},\mathbf{v}^{\mathrm{tan}}_{t,k}\}_{k\in\mathcal{K}}]$ during post-training.

\subsection{Policy Architecture and Training Stages}\label{sec:supp:method:training}

\paragraph{Teacher--student pre-training.}
We train one multi-task mimic teacher $\pi_{\mathrm{mimic}}$ following ResMimic \citep{zhao2025resmimic}. The teacher tracks 44 retargeted human--object interaction motions that span PickUp, SitStand, Push, and Carry and are collected with three training object instances: a box for PickUp, a chair for SitStand, and a large box for Push. It conditions on the current reference-motion state together with privileged body velocity, object pose, and contact information. Parallel simulation environments pair each object with its reference motion or motions, and their state distributions are combined to train this one teacher.
The student policy $\pi_{\mathrm{base}}$ is trained with DAgger-style behavior cloning on the deployment observation $o_{\mathrm{base}}$ of \cref{eq:obs_base}, in which $c^{\mathrm{root}}_t$ is the commanded planar floating-base velocity. The student receives no reference joint poses, future body motion, or contact trajectories, whereas the teacher input $o_{\mathrm{mimic}}$ in \cref{eq:bc} contains privileged full-body reference-motion information. This asymmetry lets motion-imitation supervision initialize physically feasible contact behavior while the policy interface remains motion-free at inference. All task and object rollouts are sampled jointly to update the same $\pi_{\mathrm{base}}$. This stage uses no scale or geometry randomization, which enters only in discriminative post-training.

\paragraph{Policy architecture.}
All policy variants use the same Transformer-based whole-body controller. Its input tokens encode proprioception, the root-velocity command, the interaction-type flag, and the interaction latent; its output is the whole-body action sent to the low-level controller. Pre-training, post-training, and visual distillation share this architecture and change only the training objective and the sensing modality.

\paragraph{Root-command construction and synthetic augmentation.}
Post-training draws root-velocity commands from recorded and synthetic sources. Recorded commands are finite differences of the reference root trajectory projected onto the $x$--$y$ plane. To supplement the limited locomotion diversity of the 44 collected interaction clips, synthetic samples compose coarse root and object command primitives through predefined task templates. These primitives match the command family used in evaluation, but no held-out composed evaluation sequence is reused.

\begin{table}[t]
    \centering
    \small
    \setlength{\tabcolsep}{3pt}
    \caption{\textbf{Training hyperparameters of the three stages.} The stages share one policy architecture and differ in their optimization settings.}
    \label{tab:supp_training_details}
    \begin{tabular*}{\linewidth}{@{\extracolsep{\fill}}lcc@{}}
        \toprule
        \textbf{Hyperparameter} & \textbf{Symbol} & \textbf{Value} \\
        \midrule
        \rowcolor{gray!30}
        \multicolumn{3}{l}{\textbf{Interaction Skill Pre-Training}} \\
        Learning rate & $\eta_{\mathrm{pre}}$ & $1\times10^{-3}$ \\
        Batch size & $B_{\mathrm{pre}}$ & $4096\times8$ \\
        Number of training iterations & $N_{\mathrm{pre}}$ & $24{,}000$ \\
        Optimizer & -- & Adam \\
        \midrule
        \rowcolor{gray!30}
        \multicolumn{3}{l}{\textbf{Discriminative Post-Training}} \\
        Policy learning rate & $\eta_{\mathrm{pol}}$ & $1\times10^{-3}$ \\
        Discriminator learning rate & $\eta_{\mathrm{disc}}$ & $2\times10^{-4}$ \\
        Discount factor & $\gamma$ & $0.99$ \\
        GAE lambda & $\lambda$ & $0.95$ \\
        Entropy coefficient & $\alpha_{\mathrm{ent}}$ & $5\times10^{-4}$ \\
        Number of environments & $N_{\mathrm{env}}$ & $4096\times8$ \\
        Number of environment steps & $N_{\mathrm{rl}}$ & $240{,}000$ \\
        \midrule
        \rowcolor{gray!30}
        \multicolumn{3}{l}{\textbf{Visual-Motor Distillation}} \\
        Learning rate & $\eta_{\mathrm{dist}}$ & $1\times10^{-3}$ \\
        Batch size & $B_{\mathrm{dist}}$ & $2048\times8$ \\
        Number of distillation iterations & $N_{\mathrm{dist}}$ & $120{,}000$ \\
        \bottomrule
    \end{tabular*}
\end{table}

\subsection{Discriminative Post-Training Rewards}\label{sec:supp:method:posttrain}

Post-training combines root-velocity tracking, \ac{aip} interaction regularization, a whole-body motion-style prior, and control penalties in the reward $r_t$ of \cref{eq:posttrain_reward}, with weights shared across tasks (\cref{tab:supp_reward}). The interaction reward in \cref{eq:interaction_reward} takes the least-squares form of \ac{amp} \citep{peng2021amp} and is evaluated on the velocity-augmented interaction statistic $h_t$. A separate discriminator over robot states supplies the motion-style reward. Control penalties cover action changes, joint limits, and early termination.

\begin{table}[t!]
    \centering
    \small
    \setlength{\tabcolsep}{3pt}
    \caption{\textbf{Reward terms for discriminative post-training.} In the formulations, $\mathbf{v}^{\mathrm{root},xy}_t$ is the measured planar root velocity, $c^{\mathrm{root}}_t$ its command, and $D(s)$ the motion-style discriminator over robot states.}
    \label{tab:supp_reward}
    \begin{tabular*}{\linewidth}{@{\extracolsep{\fill}}>{\raggedright\arraybackslash}p{0.20\linewidth}>{\centering\arraybackslash}p{0.28\linewidth}>{\centering\arraybackslash}p{0.08\linewidth}>{\raggedright\arraybackslash}p{0.36\linewidth}@{}}
        \toprule
        \textbf{Reward Term} & \textbf{Formulation} & \textbf{Weight} & \textbf{Description} \\
        \midrule
        \rowcolor{gray!15}
        \multicolumn{4}{l}{\textbf{Task and Style Rewards}} \\
        Root Velocity Tracking & $-\|\mathbf{v}^{\mathrm{root},xy}_t-c^{\mathrm{root}}_t\|^2$ & $1.0$ & Track the commanded planar root velocity without motion references. \\
        Interaction Style & \cref{eq:interaction_reward} & $2.0$ & Discriminator-based prior on the distance-field interaction representation. \\
        Motion Style & $1-0.25(D(s)-1)^2$ & $1.0$ & Discriminator-based motion prior for natural gait and posture. \\
        \midrule
        \rowcolor{gray!15}
        \multicolumn{4}{l}{\textbf{Regularization and Penalties}} \\
        Action Reg. & $-\|\Delta a_t\|^2$ & $5.0$ & Penalize abrupt action changes to improve control stability. \\
        Termination & Constant & $-10.0$ & Applied upon early termination, such as falls or loss of balance. \\
        Joint Limit & $n_{\mathrm{exceed}}$ & $-5.0$ & Penalize joints exceeding specified soft limits. \\
        \bottomrule
    \end{tabular*}
\end{table}

\subsection{Visual-Motor Distillation Details}\label{sec:supp:method:distill}

The vision policy replaces the privileged \ac{mocap}-derived distance-field observation with egocentric depth and observes $o_{\mathrm{vis}}$ in \cref{eq:obs_vis}, where the visual encoder $E_\phi$ is trained to predict, from the history of egocentric depth frames $S_t$, a compact interaction latent sufficient for the control policy. Distillation is DAgger-style, with the frozen \ac{mocap}-based teacher $\pi_{\mathrm{full}}$. At the visited states, the teacher is queried with its distance-field observation, and the student minimizes the action regression $\mathcal{L}_{\mathrm{distill}}$ in \cref{eq:distill}.

We apply domain randomization to camera extrinsics, depth noise, object physical parameters, and robot dynamics to improve robustness. The visual encoder is trained on depth observations from three training meshes (a small box, a large box, and a chair) and never on the real-world foam roll or soccer ball used for the held-out deployment tests.

\begin{table}[t]
    \centering
    \small
    \caption{\textbf{SitStand success (\%) across object scales with front-only and front--rear simulated depth.} A rear depth view exposes the pelvis--chair contact that the standard front-facing camera poorly observes.}
    \label{tab:supp_sitstand_fov}
    \begin{tabular*}{\linewidth}{@{\extracolsep{\fill}}lrrrrrr@{}}
        \toprule
        \textbf{Depth model} & \textbf{0.4$\times$} & \textbf{0.6$\times$} & \textbf{1.0$\times$} & \textbf{1.4$\times$} & \textbf{1.6$\times$} & \textbf{Overall} \\
        \midrule
        Front-only & 0 & 0 & 0 & 0 & 0 & 0 \\
        Front--rear & 0 & 1 & 31 & 85 & 38 & 31 \\
        \bottomrule
    \end{tabular*}
\end{table}

\section{Experimental Details}\label{sec:supp:exp}

\subsection{Tasks, Metrics, and Success Criteria}\label{sec:supp:exp:metrics}

The four skills involve different contact objectives, so each task has its own metric.

\paragraph{PickUp.}
A trial succeeds if the object is lifted above the height threshold and remains stably held for the required duration.

\paragraph{SitStand.}
A trial succeeds if the robot establishes stable pelvis--chair contact while keeping the root height within the valid range. This metric evaluates seated contact and posture stability rather than the precise reproduction of a reference motion.

\paragraph{Push.}
The hand--object contact rate $R_{\mathrm{cont}}$ is measured over the task horizon, and body-dominated contact is not counted. The metric thus measures whether the robot maintains the intended bimanual contact with the pushed object.

\paragraph{Carry and sequential multi-skill execution.}
Carry and sequential multi-skill execution are both judged by whether the manipulated object follows the commanded trajectory within a fixed tolerance. Sequential success requires completing all task instances in the sequence without resetting the robot or the environment.

\subsection{Compute Resources}\label{sec:supp:exp:compute}

All training is conducted on eight NVIDIA RTX 5090 GPUs. Training the motion-tracking mimic teacher takes approximately 24 hours, interaction skill pre-training of $\pi_{\mathrm{base}}$ approximately 1 hour, discriminative post-training with \ac{aip}-guided \ac{rl} approximately 24 hours, and visual-motor distillation approximately 4 hours. These estimates correspond to the training configurations in \cref{tab:supp_training_details}; data preprocessing, evaluation rollouts, and real-world deployment calibration may require additional wall-clock time.

\subsection{Sequential Multi-Skill Protocol}\label{sec:supp:exp:sequential}

Sequential evaluation tests the temporally extended execution of one shared low-level policy. The task sequence is specified externally, and the one-hot interaction-type flag $g_t$ provides the active interaction mode.

\paragraph{Evaluation sequence generation.}
Command sequences are synthesized only for evaluation and never serve as additional training demonstrations. For each sequence length $N$, we randomly sample $N$ task instances from PickUp, SitStand, Push, and Carry. Each task type has predefined command templates and transition rules that specify the interaction-type flag, the nominal task phase, and the condition for switching to the next task. Within each sampled task instance, the object geometry, object pose, and root-velocity command are randomized according to the task distribution. The benchmark therefore evaluates execution under randomized commanded task sequences rather than imitation of a prerecorded long-horizon motion reference.

\paragraph{Execution and success.}
During training, skill labels come from annotations of the 44 collected interaction trajectories, but the policy is never trained on full composed multi-task sequences. At evaluation, the operator or task script updates the one-hot interaction-type flag at task boundaries and provides joystick or scripted root commands, which specify the desired planar velocity of the floating base. The policy must execute each commanded skill from its current physical state, without environment resets or switching to a skill-specific policy, and a sequence counts as successful only if all $N$ task instances are completed.

This protocol is challenging because physical errors accumulate across skills. Small root-tracking drift, object displacement, unstable contact, or balance errors can shift the state distribution before the next skill begins. The sequential evaluation therefore measures the robustness of a shared low-level interaction policy under accumulated state drift.

\subsection{Baselines and Ablations}\label{sec:supp:exp:baselines}

We compare against representative motion-conditioned and motion-free-at-inference baselines, with matched task definitions and scale settings wherever their required observations and reference inputs are available. Motion-conditioned baselines require motion or pose references at inference, whereas \model uses a root command, an interaction-type flag, and geometric interaction observations. The full model, HDMI, and ResMimic share the per-axis scale-randomization interval $[0.6,1.4]$, so evaluations at $0.4\times$ and $1.6\times$ test extrapolation beyond it.

The sequential benchmark compares ablations of the same deployed policy. HDMI and ResMimic accept motion and object references, including the transitions needed for a composed sequence; PhysHSI and VisualMimic use separate policies for individual skills. These differences in input and policy structure limit direct comparison under the interaction-type flag and root-command protocol used here.

The ablations in the main paper isolate five components.
\begin{itemize}[nosep,leftmargin=*]
    \item \textbf{Ours\,-\,AIP} removes the adversarial interaction reward and the discriminator update, while keeping the distance-field policy input and the rest of the \ac{rl} reward stack unchanged.
    \item \textbf{Ours\,-\,Syn.} removes the synthetic interaction data augmentation.
    \item \textbf{Ours\,-\,Rand.} disables geometry randomization.
    \item \textbf{Ours\,-\,RL} removes \ac{rl} fine-tuning after behavior cloning.
    \item \textbf{Ours\,-\,Trans.} replaces the Transformer backbone with a \ac{mlp} backbone.
\end{itemize}

\subsection{Simulation Evaluation Protocol}\label{sec:supp:exp:sim}

For every simulation result, including the representation ablation, the additional-geometry evaluation, and the front--rear depth ablation, we evaluate each trained policy over 128 episodes per task and scale. Results are averaged over three seeds, and standard deviations, where reported, are computed over seed-level averages. The mean root-position error across the evaluated simulation motions is $0.1298\,\mathrm{m}$. Sequential trials use the all-instance completion criterion of \cref{sec:supp:exp:sequential}. Dashes in the result tables mark settings unsupported by the original method interface or unavailable under the required observation and reference inputs.

\subsection{Real-World Evaluation Protocol}\label{sec:supp:exp:real}

Real-world evaluation has qualitative and quantitative components. Qualitatively, we test all four skills (PickUp, SitStand, Push, and Carry) and sequential multi-skill execution. Quantitatively, we report ten repeated trials per setting for PickUp and SitStand, because these tasks have reliable and repeatable measurement protocols on hardware; Push, Carry, and sequential execution are reported qualitatively. External \ac{mocap} measures root-tracking accuracy for evaluation only. The vision policy is not evaluated quantitatively on SitStand, because the egocentric camera poorly observes the key back-side chair contact.

\subsection{Geometry, Object--Task Transfer, and Representation Evaluation}\label{sec:supp:exp:ood}

\Cref{tab:supp_ood_repr} compares four representations under the same training protocol on five PickUp meshes, four held out from training and one paired with a new task, at five scales. Latent \ac{df} has the highest scale-averaged success on four meshes and raw \ac{df} on the chair. Highly curved objects remain difficult for all representations; the dominant failures are hand--object slippage and, for spherical objects, rolling before stable contact is established.

\begin{table}[t]
    \centering
    \scriptsize
    \setlength{\tabcolsep}{3pt}
    \caption{\textbf{PickUp success (\%) on additional object geometries and an untrained object--task pairing.} The sphere, stool, cylinder, and torus meshes are held out from training; the chair mesh is seen in training for other interactions but not for PickUp. Each row reports success at scales from $0.4\times$ to $1.6\times$ and the mean over scales.}
    \label{tab:supp_ood_repr}
    \begin{tabular*}{\linewidth}{@{\extracolsep{\fill}}llrrrrrr@{}}
        \toprule
        \textbf{Representation} & \textbf{Geometry} & \textbf{0.4} & \textbf{0.6} & \textbf{1.0} & \textbf{1.4} & \textbf{1.6} & \textbf{Overall} \\
        \midrule
        Latent DF & Chair & 63.28 & 95.31 & 99.22 & 18.75 & 1.56 & 55.62 \\
        Latent DF & Sphere & 54.69 & 1.56 & 0.00 & 0.00 & 0.00 & 11.25 \\
        Latent DF & Stool & 48.44 & 99.22 & 100.00 & 42.19 & 47.66 & 67.50 \\
        Latent DF & Cylinder & 0.78 & 39.84 & 14.06 & 0.00 & 0.00 & 10.94 \\
        Latent DF & Torus & 0.00 & 0.00 & 22.66 & 0.00 & 1.56 & 4.84 \\
        \midrule
        BBox & Chair & 16.41 & 17.97 & 51.56 & 21.88 & 4.69 & 22.50 \\
        BBox & Sphere & 0.00 & 0.78 & 0.00 & 0.00 & 0.00 & 0.16 \\
        BBox & Stool & 46.09 & 27.34 & 7.81 & 2.34 & 1.56 & 17.03 \\
        BBox & Cylinder & 1.56 & 12.50 & 2.34 & 0.00 & 0.00 & 3.28 \\
        BBox & Torus & 0.00 & 0.00 & 1.56 & 0.00 & 3.91 & 1.09 \\
        \midrule
        Voxel & Chair & 68.75 & 96.88 & 96.88 & 21.88 & 2.34 & 57.34 \\
        Voxel & Sphere & 0.00 & 0.00 & 0.00 & 0.00 & 0.00 & 0.00 \\
        Voxel & Stool & 46.09 & 27.34 & 91.41 & 14.84 & 5.47 & 37.03 \\
        Voxel & Cylinder & 0.78 & 17.19 & 0.00 & 0.00 & 0.00 & 3.59 \\
        Voxel & Torus & 0.00 & 0.00 & 3.12 & 3.91 & 0.00 & 1.41 \\
        \midrule
        Raw DF & Chair & 92.97 & 91.41 & 92.97 & 18.75 & 9.38 & 61.09 \\
        Raw DF & Sphere & 3.91 & 0.00 & 0.00 & 0.00 & 0.00 & 0.78 \\
        Raw DF & Stool & 78.91 & 70.31 & 89.06 & 27.34 & 21.88 & 57.50 \\
        Raw DF & Cylinder & 0.00 & 27.34 & 0.00 & 0.00 & 0.00 & 5.47 \\
        Raw DF & Torus & 0.00 & 0.00 & 9.38 & 3.91 & 0.00 & 2.66 \\
        \bottomrule
    \end{tabular*}
\end{table}

We also tested a second new object--task pairing, the Push command on the chair, and the policy pushed the chair in simulation and on hardware. Hardware attempts to pick up the chair were limited by motor overheating under the chair's weight.

\section{Additional Analysis}\label{sec:supp:analysis}

\subsection{Representation Similarity Across Tasks and Scales}\label{sec:supp:analysis:latent}

The latent similarities in \cref{tab:supp_latent_similarity} use the \ac{vae} encoder mean $\mu_\phi(I_t)$ as the interaction latent instead of a sample from the posterior. Latents are collected from evaluation rollouts of the full \model policy, with ten rollout trajectories for each task--scale pair. For each trajectory, we extract the segment within a 5-second window around the main contact event, average its latent vectors into one trajectory-level interaction representation, and $\ell_2$-normalize this representation before computing cosine similarities.

Each same-scale diagonal entry averages the pairwise cosine similarities between different trajectories of the same task at the nominal scale $1.0\times$, excluding self-pairs. Each value in parentheses averages the pairwise similarities between trajectory-level representations of the same task over all pairs of different scales. Off-diagonal entries average the pairwise similarities between trajectory-level representations of different tasks at the nominal scale $1.0\times$. Each same-scale entry thus uses ten trajectories per task at the nominal scale, and each cross-scale entry uses ten trajectories for every task--scale pair.

\begin{table}[t]
    \centering
    \small
    \setlength{\tabcolsep}{6pt}
    \caption{\textbf{Cosine similarity of interaction latents across tasks and scales.} Each trajectory-level latent is the \acs{vae} encoder mean averaged over a 5-second window around the main contact event. Diagonal entries report same-task similarity at the nominal scale $1.0\times$, values in parentheses same-task similarity averaged over all pairs of different scales, and off-diagonal entries cross-task similarity at $1.0\times$.}
    \label{tab:supp_latent_similarity}
    \begin{tabular*}{\linewidth}{@{\extracolsep{\fill}}lcccc@{}}
        \toprule
        & Carry & PickUp & Push & SitStand \\
        \midrule
        Carry & 0.51 (0.24) & 0.41 & -0.02 & -0.04 \\
        PickUp & 0.41 & 0.55 (0.28) & -0.03 & -0.04 \\
        Push & -0.02 & -0.03 & 0.39 (0.08) & -0.06 \\
        SitStand & -0.04 & -0.04 & -0.06 & 0.48 (0.22) \\
        \bottomrule
    \end{tabular*}
\end{table}

Interaction latents retain higher intra-task similarity than unrelated inter-task pairs, and PickUp and Carry share a relatively high mutual similarity because both involve sustained bimanual object interaction. Same-task similarity is lower across scales than at the nominal scale for every task, and Push shows the weakest cross-scale similarity, which suggests that pushing behavior is more sensitive to object scale and contact timing than PickUp or Carry. The latent is therefore not strictly scale-invariant; it remains task-structured while preserving scale-dependent changes in interaction.

\subsection{\texorpdfstring{\acs{aip}}{AIP} Trade-Off Analysis}\label{sec:supp:analysis:aip}

Removing \ac{aip} leaves the \ac{df} policy observation and the other \ac{rl} rewards unchanged. Carry success then falls to zero, and sequential success falls to 5.2\% at $N=5$ and to zero for $N\ge10$. The Push contact rate and the SitStand success at all three reported scales instead increase. This pattern suggests that matching interaction statistics is particularly useful for maintaining coordinated contact across multiple phases, whereas optimizing the remaining rewards can favor individual contact metrics. Motion-conditioned baselines likewise obtain higher scores in some settings where the geometry is compatible with their references (\cref{tab:unified_sim}).

\paragraph{Discriminator input and numerical stability.}
The raw-\ac{df} representation ablation changes the policy observation but retains the encoded geometric input to \ac{aip}, and this configuration completes post-training. Replacing the latent input of the discriminator with raw \ac{df} features instead produced NaNs and prevented post-training from completing in our trials. These observations motivated retaining the latent interaction statistic in the reported representation comparison.

\subsection{Vision Policy Observability Analysis}\label{sec:supp:analysis:vision}

The vision policy replaces privileged distance-field observations with a depth-predicted interaction latent. This replacement makes deployment possible without \ac{mocap} but introduces partial observability. The limitation is most visible in SitStand, where the key pelvis--chair contact occurs behind the robot and is poorly observed by the front-facing egocentric camera.

The front--rear depth ablation in \cref{tab:supp_sitstand_fov} tests the role of visual coverage. The front-only model achieves 0\% success at every scale, whereas adding a rear view raises the scale-averaged success to 31\%. The remaining gap to privileged \ac{df} sensing may reflect residual occlusion and errors in estimating the interaction latent. The standard front-only hardware vision evaluation therefore covers PickUp, and quantitative SitStand trials use the \ac{mocap}-based policy.

The low sequential success of the vision policy is partly associated with the same observability limitation. When sequences involving SitStand are excluded, the vision policy reaches 36\% success, which shows that the depth-predicted latent can support sequential execution when the relevant contact surfaces are visible. Improved camera placement, rear-facing sensors, or multi-view onboard perception could extend visual coverage to these interactions.

\subsection{Failure Modes and Limitations}\label{sec:supp:exp:failure}

\begin{figure}[t!]
    \centering
    \includegraphics[width=\linewidth]{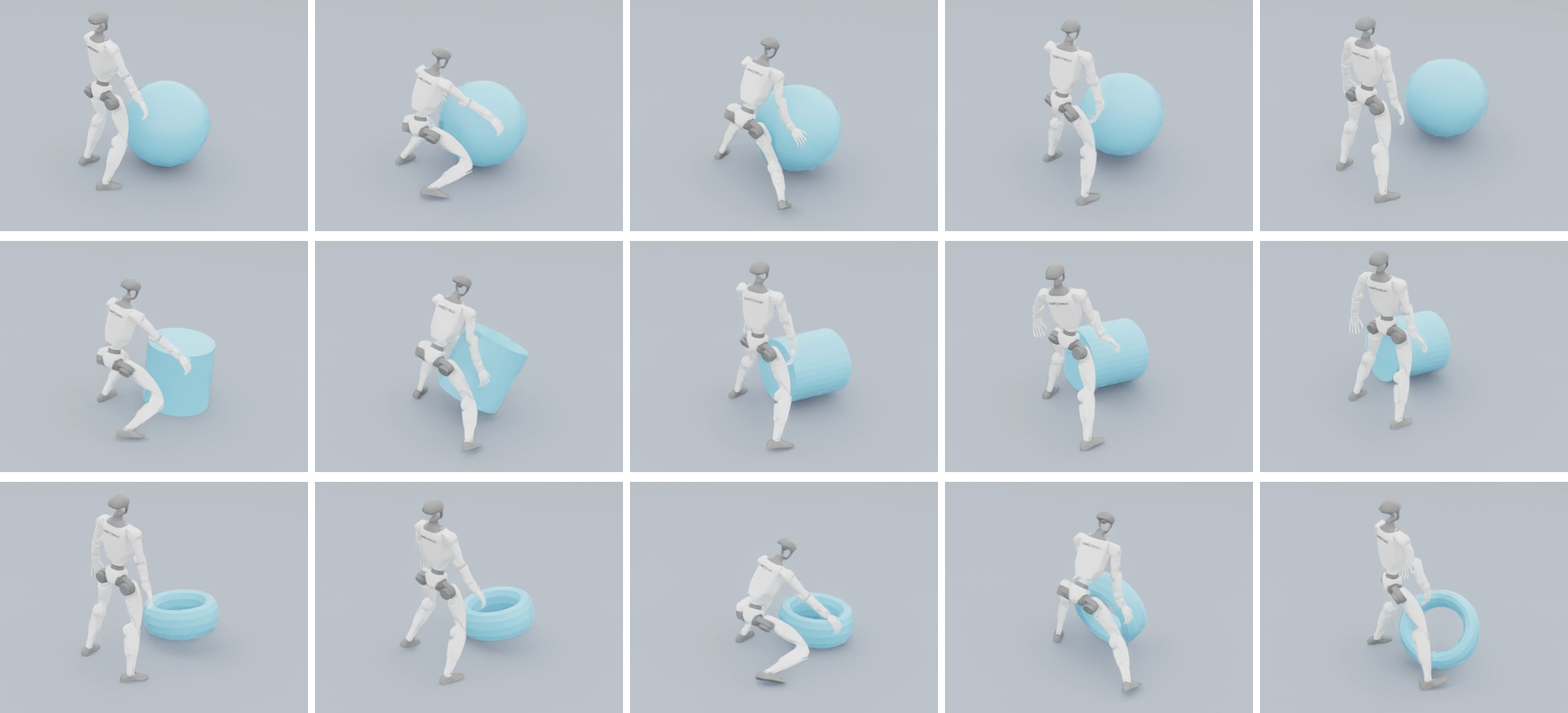}
    \caption{\textbf{Representative failures on curved objects.} Unsuccessful interactions with a sphere, a cylinder, and a torus.}
    \label{fig:supp:failure_curved_surface}
\end{figure}

\begin{figure}[t!]
    \centering
    \includegraphics[width=\linewidth]{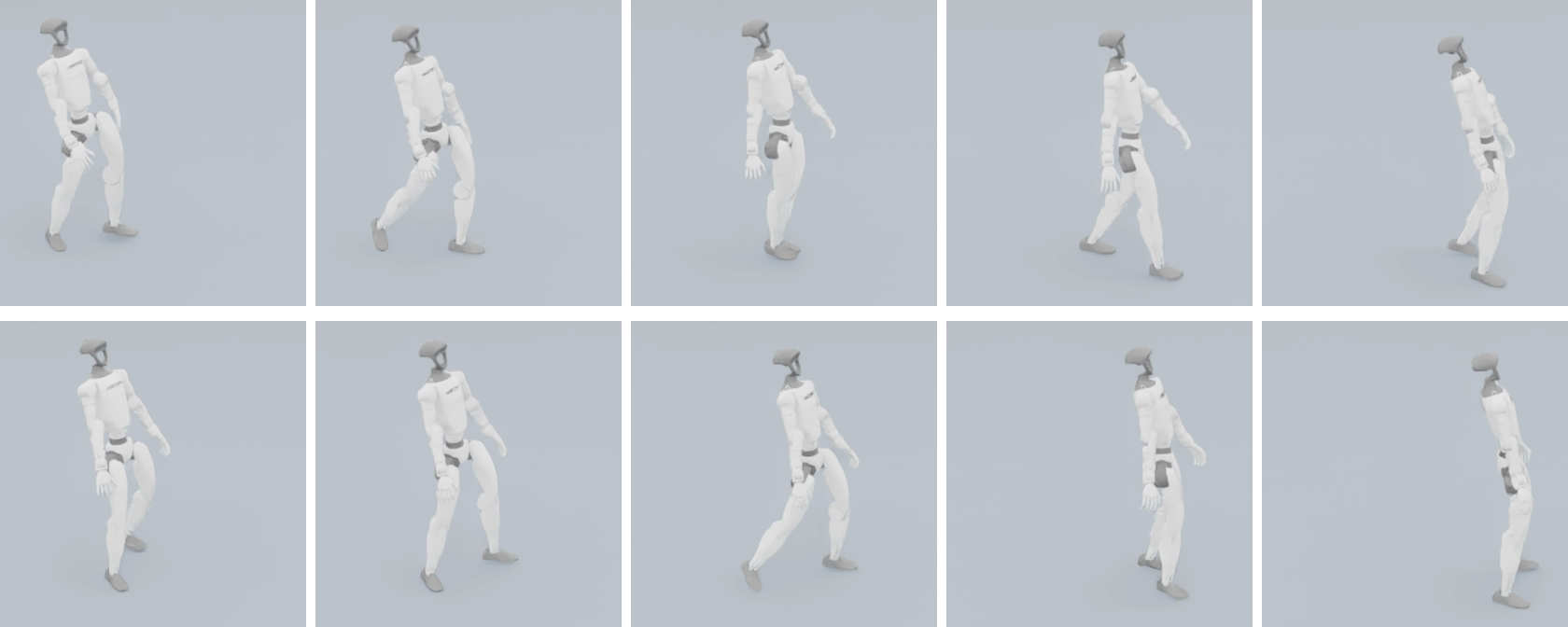}
    \caption{\textbf{Gait of the mimic teacher and the post-trained policy.} Matched rollouts of $\pi_{\mathrm{mimic}}$ (top) and $\pi_{\mathrm{full}}$ (bottom) under the same configuration.}
    \label{fig:supp:stage_comparison}
\end{figure}

Most observed failures occur when an interaction requires timing, force precision, or visibility beyond what the current sensing and per-link geometric interface provide. Sensing latency in the \ac{mocap} or vision pipeline can delay contact during fast motions. Mismatched contact force can also cause failure, because excessive force may destabilize the object or the robot, whereas insufficient force can lead to missed grasps, weak pushing, or failed carrying. Objects that require precise force balance or fine contact placement remain challenging, since the per-link \ac{df} representation captures coarse proximity and surface direction rather than a detailed pressure distribution. For vision-based deployment, failures occur mainly when task-relevant contacts are occluded or outside the egocentric camera view, such as the back-side chair contact in SitStand. \Cref{fig:supp:failure_curved_surface} shows representative failures on objects with curved surfaces.

Hardware motion also shows small steps, low foot clearance, and occasional jitter. Slight out-toeing and low-clearance gait patterns are already present in the retargeted references and in the mimic teacher, and post-training adds a modest degradation in gait quality. The interaction and whole-body style objectives may produce competing corrections under randomized geometry. \Cref{fig:supp:stage_comparison} compares matched rollouts of the mimic teacher and the post-trained policy. Sensing errors and latency further affect motion on hardware. Temporary errors in the \ac{mocap} root pose occur when the foam roll occludes the cameras, and the Python deployment stack adds roughly 5 to 15\,ms of variable control latency.

Future improvements include reward balancing and temporal regularization for smoother motion, an optimized C++/ONNX control stack, force-aware feedback, denser geometric queries, and wider visual coverage.

\newpage
\section*{NeurIPS Paper Checklist}

\begin{enumerate}

\item {\bf Claims}
    \item[] Question: Do the main claims made in the abstract and introduction accurately reflect the paper's contributions and scope?
    \item[] Answer: \answerYes{}
    \item[] Justification: The main claims made in the abstract and introduction accurately reflect the paper’s contributions.
    \begin{itemize}
        \item The answer \answerNA{} means that the abstract and introduction do not include the claims made in the paper.
        \item The abstract and/or introduction should clearly state the claims made, including the contributions made in the paper and important assumptions and limitations. A \answerNo{} or \answerNA{} answer to this question will not be perceived well by the reviewers. 
        \item The claims made should match theoretical and experimental results, and reflect how much the results can be expected to generalize to other settings. 
        \item It is fine to include aspirational goals as motivation as long as it is clear that these goals are not attained by the paper. 
    \end{itemize}

\item {\bf Limitations}
    \item[] Question: Does the paper discuss the limitations of the work performed by the authors?
    \item[] Answer: \answerYes{}
    \item[] Justification: See \cref{sec:supp:exp:failure} for the limitations.

\item {\bf Theory assumptions and proofs}
    \item[] Question: For each theoretical result, does the paper provide the full set of assumptions and a complete (and correct) proof?
    \item[] Answer: \answerNA{} 
    \item[] Justification: The paper does not include theoretical results.

\item {\bf Experimental result reproducibility}
    \item[] Question: Does the paper fully disclose all the information needed to reproduce the main experimental results of the paper to the extent that it affects the main claims and/or conclusions of the paper (regardless of whether the code and data are provided or not)?
    \item[] Answer: \answerYes{} 
    \item[] Justification: We have included necessary details in the paper and the supplementary material. The training code, data, checkpoint, and deployment system will be released on our project website.

\item {\bf Open access to data and code}
    \item[] Question: Does the paper provide open access to the data and code, with sufficient instructions to faithfully reproduce the main experimental results, as described in supplemental material?
    \item[] Answer: \answerYes{} 
    \item[] Justification:  We have included necessary details in the paper and the supplementary material. The training code, data, checkpoint, and deployment system will be released on our project website.

\item {\bf Experimental setting/details}
    \item[] Question: Does the paper specify all the training and test details (e.g., data splits, hyperparameters, how they were chosen, type of optimizer) necessary to understand the results?
    \item[] Answer: \answerYes{} 
    \item[] Justification: Yes. We have included these details in the paper.

\item {\bf Experiment statistical significance}
    \item[] Question: Does the paper report error bars suitably and correctly defined or other appropriate information about the statistical significance of the experiments?
    \item[] Answer: \answerYes{} 
    \item[] Justification: Yes, we report them in our experiment section~\cref{sec:exp} and \cref{sec:supp:exp}.

\item {\bf Experiments compute resources}
    \item[] Question: For each experiment, does the paper provide sufficient information on the computer resources (type of compute workers, memory, time of execution) needed to reproduce the experiments?
    \item[] Answer: \answerYes{} 
    \item[] Justification: We have included these details in~\cref{sec:supp:exp:compute}.
    
\item {\bf Code of ethics}
    \item[] Question: Does the research conducted in the paper conform, in every respect, with the NeurIPS Code of Ethics \url{https://neurips.cc/public/EthicsGuidelines}?
    \item[] Answer: \answerYes{} 
    \item[] Justification: The research conducted in the paper conforms with the NeurIPS Code of
Ethics.

\item {\bf Broader impacts}
    \item[] Question: Does the paper discuss both potential positive societal impacts and negative societal impacts of the work performed?
    \item[] Answer: \answerNA{} 
    \item[] Justification:  There is no societal impact of the work.
    
\item {\bf Safeguards}
    \item[] Question: Does the paper describe safeguards that have been put in place for responsible release of data or models that have a high risk for misuse (e.g., pre-trained language models, image generators, or scraped datasets)?
    \item[] Answer: \answerNA{} 
    \item[] Justification: The paper poses no such risks.

\item {\bf Licenses for existing assets}
    \item[] Question: Are the creators or original owners of assets (e.g., code, data, models), used in the paper, properly credited and are the license and terms of use explicitly mentioned and properly respected?
    \item[] Answer: \answerYes{} 
    \item[] Justification: The assets are properly credited and the license are mentioned and respected.

\item {\bf New assets}
    \item[] Question: Are new assets introduced in the paper well documented and is the documentation provided alongside the assets?
    \item[] Answer: \answerNA{} 
    \item[] Justification: The paper does not introduce or release a new public dataset, benchmark, or model asset. The supplementary material only provides implementation details, experimental protocols, and qualitative results needed to support reproducibility.

\item {\bf Crowdsourcing and research with human subjects}
    \item[] Question: For crowdsourcing experiments and research with human subjects, does the paper include the full text of instructions given to participants and screenshots, if applicable, as well as details about compensation (if any)? 
    \item[] Answer: \answerNA{}
    \item[] Justification: The paper does not involve crowdsourcing or human-subjects experiments. Human operators are only used to provide high-level robot commands, such as joystick root commands and interaction-type flags, during system deployment and evaluation; the study does not collect or analyze human-subject data.

\item {\bf Institutional review board (IRB) approvals or equivalent for research with human subjects}
    \item[] Question: Does the paper describe potential risks incurred by study participants, whether such risks were disclosed to the subjects, and whether Institutional Review Board (IRB) approvals (or an equivalent approval/review based on the requirements of your country or institution) were obtained?
    \item[] Answer: \answerNA{} 
    \item[] Justification: The paper does not involve crowdsourcing or human-subjects experiments.

\item {\bf Declaration of LLM usage}
    \item[] Question: Does the paper describe the usage of LLMs if it is an important, original, or non-standard component of the core methods in this research? Note that if the LLM is used only for writing, editing, or formatting purposes and does \emph{not} impact the core methodology, scientific rigor, or originality of the research, declaration is not required.
    \item[] Answer: \answerNA{} 
    \item[] Justification: LLMs were not used as a component of the core methodology or experimental design. Their use was limited to basic text refinement, such as spell checking and grammar suggestions.

\end{enumerate}

\end{document}